\documentclass[10pt,journal,compsoc]{IEEEtran}

\usepackage{threeparttable}

\ifCLASSOPTIONcompsoc
  \usepackage[nocompress]{cite}
\else
  \usepackage{cite}
\fi
\ifCLASSINFOpdf
\else
\fi
\usepackage{booktabs}
\usepackage{graphicx}
\usepackage{subfigure}
\usepackage{caption}
\usepackage{color}
\usepackage{xcolor}
\usepackage{multirow}
\usepackage{makecell}
\usepackage{array}
\usepackage{subfigure}
\usepackage{threeparttable}
\begin{document}

%
\title{DFME: A New Benchmark for Dynamic Facial Micro-expression Recognition}
%
%
%
%

\author{
        Sirui~Zhao,
        Huaying~Tang,
        Xinglong~Mao,
        Shifeng~Liu,
        Yiming~Zhang,
        Hao Wang,
        Tong~Xu,~\IEEEmembership{Member,~IEEE,}
        and~Enhong~Chen,~\IEEEmembership{Fellow,~IEEE}
\IEEEcompsocitemizethanks{\IEEEcompsocthanksitem Sirui Zhao is with the School of Computer Science and Technology, University of Science and Technology of China, Hefei, Anhui 230027, China, and also with the School of Computer Science and Technology, Southwest University of Science and Technology, Mianyang 621010, China.\protect\\
E-mail: sirui@mail.ustc.edu.cn

\IEEEcompsocthanksitem Huaying Tang is with the School of Computer Science and Technology, University of Science and Technology of China, Hefei, Anhui 230027, China.\protect\\
E-mail: iamthy@mail.ustc.edu.cn
\IEEEcompsocthanksitem Xinglong Mao, Shifeng Liu, Yiming Zhang, Hao Wang, Tong Xu and Enhong Chen are with School of Data Science, University of Science and Technology of China, Hefei, Anhui 230027, China.\protect\\
E-mail: \{maoxl, lsf0619, ymzhang21\}@mail.ustc.edu.cn, \\\{wanghao3, tongxu, cheneh\}@ustc.edu.cn\protect
}
\thanks{This work was fully supported by the National Natural Science Foundation of China (No.61727809, 62072423), the Young Scientists Fund of the Natural Science Foundation of Sichuan Province (No.2023NSFSC1402). (\textit{Sirui Zhao, Huaying Tang, Xinglong Mao and Shifeng Liu contributed equally.\\ Corresponding authors: Enhong Chen and Tong Xu}).} 
\thanks{Manuscript received December xx, xx; revised xx xx, xx.}}

%
%

\markboth{Journal of \LaTeX\ Class Files,~Vol.~14, No.~8, August~2022}%
{Sirui Zhao \MakeLowercase{\textit{et al.}}: Bare Advanced Demo of IEEEtran.cls for IEEE Computer Society Journals}
%



\IEEEtitleabstractindextext{%
\begin{abstract}
One of the most important subconscious reactions, micro-expression~(ME), is a spontaneous, subtle, and transient facial expression that reveals human beings' genuine emotion. Therefore, automatically recognizing ME~(MER) is becoming increasingly crucial in the field of affective computing, providing essential technical support for lie detection, clinical psychological diagnosis, and public safety. However, the ME data scarcity has severely hindered the development of advanced data-driven MER models. Despite the recent efforts by several spontaneous ME databases to alleviate this problem, there is still a lack of sufficient data. Hence, in this paper, we overcome the ME data scarcity problem by collecting and annotating a dynamic spontaneous ME database with the largest current ME data scale called DFME (Dynamic Facial Micro-expressions). Specifically, the DFME database contains 7,526 well-labeled ME videos spanning multiple high frame rates, elicited by 671 participants and annotated by more than 20 professional annotators over three years. Furthermore, we comprehensively verify the created DFME, including using influential spatiotemporal video feature learning models and MER models as baselines, and conduct emotion classification and ME action unit classification experiments. The experimental results demonstrate that the DFME database can facilitate research in automatic MER, and provide a new benchmark for this field. \textit{DFME} will be published via \textit{https://mea-lab-421.github.io}.
\end{abstract}

\begin{IEEEkeywords}
Emotion recognition, facial micro-expression, facial action units, micro-expression recognition, databases 
\end{IEEEkeywords}}

\maketitle

\IEEEdisplaynontitleabstractindextext

%
\IEEEpeerreviewmaketitle
\ifCLASSOPTIONcompsoc
\IEEEraisesectionheading{\section{Introduction}\label{sec:introduction}}
\else
\section{Introduction}
\label{sec:introduction}
\fi
\IEEEPARstart{F}{acial} expression is one of the important channels for human beings to transmit emotional signals, accounting for 55\% of our daily communication~\cite{mehrabian2017communication,Zhao2023FacialMA}. As a particular facial expression, micro-expression~(ME) typically refers to the spontaneous and subtle facial movements that appear instantaneously when individuals attempt to conceal or suppress their genuine emotions under pressure in a high-risk situation. The first discovery of the ME phenomenon can be traced back to 1966. When scanning motion picture films of psychotherapy hours, Haggard et al. observed short-lived, imperceptible facial expressions and defined them as micro-momentary expressions~\cite{haggard1966micromomentary}. Later, in 1969, Ekman et al. also discovered ME while studying films of depressed patients. Specifically, when a video of a depressed patient was shown at a slow speed, although the patient appeared happy most of the time, there was a very brief agony look, lasting only 1/12 of a second, which revealed that the patient was trying to hide her strong negative emotions from her doctor~\cite{ekman1969nonverbal}. This finding illustrates that MEs can effectively reveal the genuine emotions of individuals, so recognizing MEs can provide essential technical support in lie detection, clinical psychological diagnosis, and public safety~\cite{porter2008reading,ekman2009telling,weinberger2010intent,hunter2020emotional}.

\begin{figure*}[h]
    \centering
    \subfigure[An example of MaE with ``Happiness'' emotion.]{\includegraphics[scale=0.55]{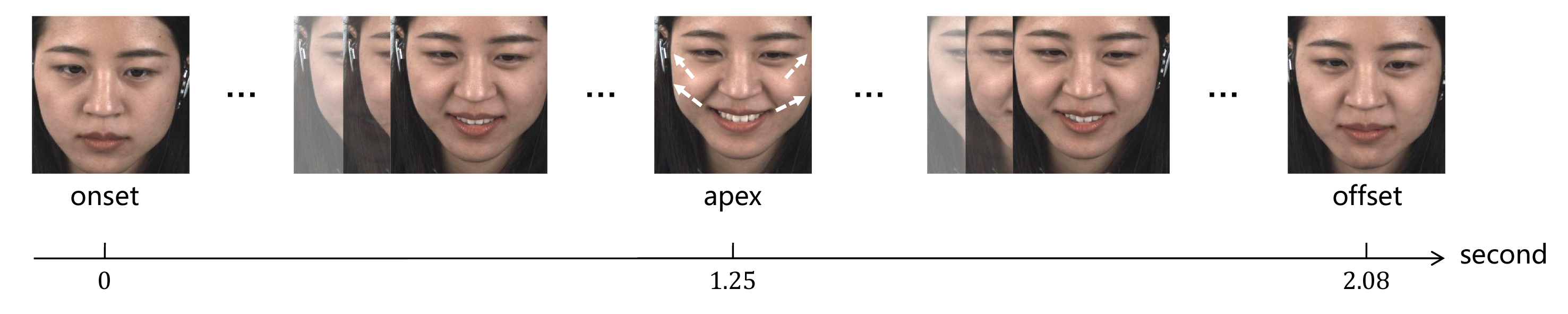}}
    \subfigure[An example of ME with ``Happiness'' emotion.]{\includegraphics[scale=0.55]{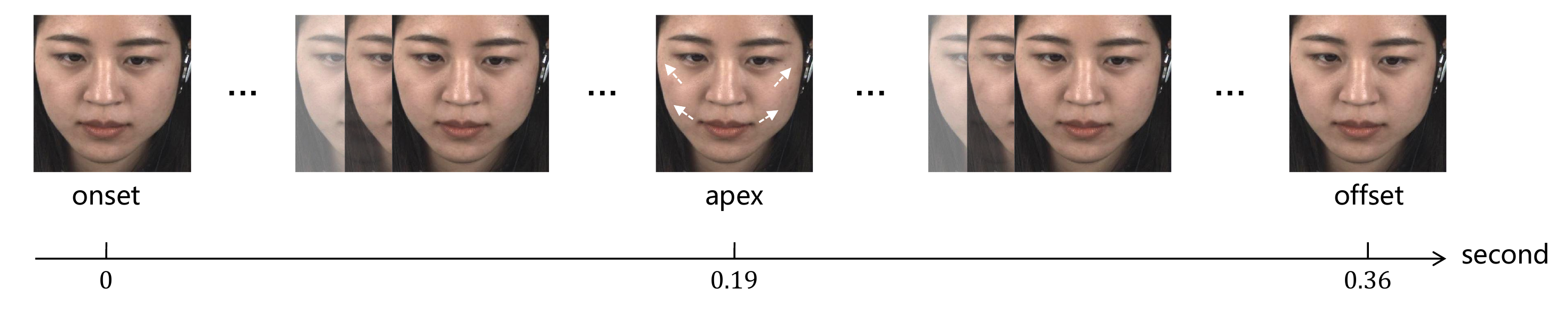}}
    \caption{Examples of MaE and ME from the same person with a timeline in seconds, both belong to the ``Happiness'' emotion category. Noteworthy, the onset frame and the offset frame denote the start and end time of an expression respectively, and the apex frame represents the moment when an expression changes most dramatically. White arrows on the face of the apex frame indicate the general directions of facial movements, and the longer and thicker the arrows, the greater the intensity of facial movements.}
    \label{ma_mi}
\end{figure*}

Essentially, ME is a subconscious reaction that an individual's willpower cannot control\cite{microreactions,Lombardi2017PsychologicalSD, wang2021single,Zhao2023FacialMA}. Compared with ordinary facial expressions also called macro-expression~(MaE), ME is hidden and has a short duration (less than 0.5s~\cite{ekman2009telling}), partial movement, and low movement intensity, so it is challenging to recognize MEs accurately. Fig.~\ref{ma_mi} compares an ME and an MaE with the same emotion category, highlighting that ME is more difficult to distinguish than MaE.

In recent years, research on ME recognition (MER) has made significant progress in interdisciplinary fields, ranging from psychology to computer science. Initially, studies~\cite{ekman2003micro,Russell2006APS,Endres2009MicroexpressionRT,frank2009see} on MER merely focused on manual analysis in psychology. However, the manual analysis relies on expert experience, which is time-consuming, labor-intensive, and has low recognition accuracy. Therefore, it is urgent to use the computers' powerful perception and computing power for automatic MER. Many efforts in the fields of computer vision and affective computing have been devoted to automatic MER. For example, in order to extract the spatial-temporal feature of MEs, Pfister et al.~\cite{pfister2011recognising} introduced a local binary pattern from three orthogonal planes~(LBP-TOP)~\cite{zhao2007dynamic} for MER. Wang et al.~\cite{wang2015lbp} proposed LBP with six intersection points (LBP-SIP) to reduce the redundant information in LBP-TOP. Huang et al.~\cite{huang2016spontaneous} used spatiotemporal completed local quantized patterns (STCLQP) for ME feature extraction. Liu et al.~\cite{liu2015main} proposed Main Directional Mean Optical Flow~(MDMO) to model the movement features of MEs. Besides the above-mentioned hand-crafted methods, many autonomous feature learning methods based on deep neural networks have also been proposed and become the mainstream method of current MER. To name a few, Wang et al.~\cite{wang2018micro} proposed Transferring Long-term Convolutional Neural Network~(TLCNN). Zhao et al.~\cite{zhao2021two} developed a novel two-stage learning (i.e., prior and target learning) method based on a siamese 3D convolutional neural network for MER. To address the occlusion problem in MER, Mao et al.~\cite{9854172} proposed a Region-inspired Relation Reasoning Network~(RRRN) to model the relationship between different facial regions. However, due to the lack of support for abundant well-labeled ME data, the recognition accuracy and robustness of these methods are challenging to meet the needs of actual scenarios~\cite{ben2021video,Zhao2023FacialMA,Li2021DeepLF,Xia2020LearningFM}. Therefore, it is urgent to build a large-scale ME database.

Although researchers have published several spontaneous ME databases, such as SMIC~\cite{li2013spontaneous}, CASME \uppercase\expandafter{\romannumeral2}~\cite{yan2014casme}, SAMM~\cite{davison2016samm}, MMEW~\cite{ben2021video} and CAS(ME)$^3$~\cite{li2022cas}, their limited sample sizes still cannot completely meet the needs of deep MER models for large-scale ME samples. It should be noted that building a large-scale spontaneous ME database involves many challenges, mainly from three aspects: First, it is challenging to induce MEs because they are facial movements disclosed after an individual attempts to suppress them. Second, it is difficult to label and distinguish ME fragments because the movement of ME is weak and fast, which is hard for the naked eye to perceive. Third, due to the short duration of MEs, high-speed cameras are often required to collect them. However, the video data stream captured by high-speed cameras is often massive and redundant, so labeling ME clips is extremely time-consuming and labor-intensive. Meanwhile, some studies\cite{MEGC2022,xu2021famgan,xu2021famgan}, have also attempted to expand the existing ME sample size by using generative methods. Specifically, by utilizing generative models and existing ME videos, researchers induce deformations in easily obtainable facial images to create new ME videos. For instance, Xu et al.~\cite{xu2021famgan} introduced Generative Adversarial Network based on fine-grained facial action units~(AUs) modulation to generate ME sequences. Zhao et al.~\cite{zhao2022fine} combined a motion estimation network based on Thin-Plate Spline with a generation network constrained by relative AUs to accurately generate fine-grained MEs. Though ME generation techniques partially alleviate the scarcity of existing ME samples, the quality of generated ME videos and the diversity of movements still cannot replace authentically captured MEs.

In order to overcome the ME data shortage bottleneck, this paper collects and annotates the currently largest ME database called DFME~(Dynamic Facial Micro-expressions) to advance the development of MER. Specifically, the proposed DFME database contains 7,526 well-labeled ME videos spanning multiple high frame rates, i.e., \textit{200fps, 300fps, 500fps}, primarily annotated with seven discrete emotion labels, i.e., \textit{happiness, anger, contempt, disgust, fear, sadness} and \textit{surprise}, along with 24 facial AU labels listed in Table~\ref{tab_AU}. In particular, our ME videos are produced by conducting video emotional stimulation on 671 participants while suppressing emotions as much as possible, and are repeatedly annotated by more than 20 professional annotators over three years. The well-labeled ME samples with cropped regions of the facial images will be publicly available to the MER research community. In addition, we comprehensively validate the created DFME database, including developing a new database validation strategy and reproducing several influential spatiotemporal video feature learning models and MER models to conduct basic emotion classification and ME action unit classification experiments. The experimental results demonstrate that the DFME database can facilitate research in automatic MER, and provide a new benchmark for this field. 

The rest of this paper is organized as follows. First, we summarize currently existing ME databases and review related work on MER in the next section. In section 3, we elaborate on the building details and statistical properties of our DFME database. Then the comprehensive database evaluation is developed and discussed in Section 4. Finally, research conclusions and future work are addressed in Section 5.

\section{Related Work}

In this section, we first review the existing public spontaneous ME databases related to MER. Then, we mainly summarize some representative MER studies based on deep learning technologies.
\begin{table*}[htbp]
    \caption{Statistical Information of Current Spontaneous ME Databases}\label{tab_ExistDatabases}
    \centering
    \scriptsize
    \renewcommand{\arraystretch}{1.5}
    \begin{threeparttable}
        \begin{tabular}{cccccccccc} 
        \toprule
        \multicolumn{2}{c}{\multirow{2}*{ME Databases}}&\multicolumn{3}{c}{Participants}&\multicolumn{3}{c}{Samples of MEs}&\multicolumn{2}{c}{Annotation Labels}\\\cmidrule(r){3-5} \cmidrule(r){6-8} \cmidrule(r){9-10}
        &&Number&\makecell[c]{Gender\\ \tiny{(Male/Female)}}&Age&Number&Frame Rate&Resolution&Emotion&FACS AU \\\midrule
         &HS&16&&&164&100&640$\times$480&Pos (51) Neg (70) Sur (43)&\\\cmidrule(r){2-3} \cmidrule(r){6-9} 
        SMIC&VIS&8&10/6&\makecell[c]{Range: 22-34\\ Mean=28.1}&71&25&640$\times$480&Pos (28) Neg (23) Sur (20)&No \\\cmidrule(r){2-3} \cmidrule(r){6-9}
        &NIR&8&&&71&25&640$\times$480&Pos (28) Neg (23) Sur (20)& \\\cmidrule(r){1-10}
        \multicolumn{2}{c}{CASME}& 35& 22/13& Mean=22.03 & 195& 60& \makecell[c]{640$\times$480\\ 1280$\times$720}&\makecell[c]{Amu (5) Dis (88) Fear (2) \\Con (3) Sad (6) Tense (28)\\ Sur (20) Rep (40)}& Yes\\\cmidrule(r){1-10}
        \multicolumn{2}{c}{CASME \uppercase\expandafter{\romannumeral2}}& 35& /& Mean=22.03& 247& 200& 640$\times$480&\makecell[c]{Hap (33) Dis (60) Sur (25) \\Rep (27) Oth (102)}& Yes\\\cmidrule(r){1-10}
        \multicolumn{2}{c}{CAS(ME)$^{2}$}& 22& 9/13& \makecell[c]{Range: 19-26\\Mean=22.59}& 57& 30& 640$\times$480&\makecell[c]{Pos (8) Neg (21) Sur (9) \\Oth (19)}& Yes\\\cmidrule(r){1-10}
        \multicolumn{2}{c}{SAMM}& 32& 16/16& \makecell[c]{Range: 19-57\\Mean=33.24}& 159& 200& 2040$\times$1088&\makecell[c]{Hap (24) Dis (8) Fear (7) \\Ang (20) Sur (13) Sad (3)\\ Oth (84)}& Yes\\\cmidrule(r){1-10}
        \multicolumn{2}{c}{MEVIEW}& 16& /&/& 29& 30& 1280$\times$720&\makecell[c]{Hap (5) Dis (1) Fear (3) \\Ang (1) Sur (8) Con(4)\\ Unc (7)}& Yes\\\cmidrule(r){1-10}
        \multicolumn{2}{c}{MMEW}& 36& /&Mean=22.35& 300& 90& 1920$\times$1080&\makecell[c]{Hap (36) Dis (72) Fear (16) \\Ang (8) Sur (89) Sad (13) \\Oth (66)}& Yes\\\cmidrule(r){1-10}
        \multirow{3}*{CAS(ME)$^{3}$}&PART A&100& 50/50&/ &943&30&1280$\times$720&\makecell[c]{Hap (64) Dis (281) Fear (93) \\Ang (70) Sur (201) Sad (64)\\ Oth (170)}&\multirow{3}{*}{Yes}\\\cmidrule(r){2-9}
        &PART C&31& 9/22&Mean=23.5 &166&30&1280$\times$720&\makecell[c]{Pos (16) Neg(99) Sur (30) \\Oth (20)}& \\\cmidrule(r){1-10}
        \multirow{5}*{4DME}& DI4D&\multirow{5}*{65}& \multirow{5}*{38/27}&\multirow{5}{*}{\makecell[c]{Range: 22-57\\ Mean=27.8}} &267&60&1200$\times$1600&\multirow{5}{*}{\makecell[c]{Pos (34) Neg (127) Sur (30) \\Rep (6) PosSur (13) NegSur (8) \\RepSur (3) PosRep(8) \\NegRep(7) Oth(31)}}&\multirow{5}{*}{Yes}\\\cmidrule(r){2-2} \cmidrule(r){6-8}
        &Grayscale&&&&267&60&640$\times$480&& \\\cmidrule(r){2-2} \cmidrule(r){6-8}
        &RGB&&&&267&30&640$\times$480&& \\\cmidrule(r){2-2} \cmidrule(r){6-8}
        &Depth&&&&267&30&640$\times$480&& \\\cmidrule(r){2-2} \cmidrule(r){1-10}
         &PART A&72& 31/41& &\textbf{1118}&\textbf{500}&1024$\times$768&\makecell[c]{Hap (111) Dis (321) Fear (143) \\Ang (97) Con (77) Sur (187) \\Sad (142) Oth (40)}&\\\cmidrule(r){2-4} \cmidrule(r){6-9}
        \textbf{DFME}&PART B&92& 61/31&\makecell[c]{Range: 17-40\\ Mean=22.43}&\textbf{969}&\textbf{300}&1024$\times$768&\makecell[c]{Hap (78) Dis (406) Fear (115) \\Ang (56) Con (45) Sur (143) \\Sad (119) Oth (7)}&Yes \\\cmidrule(r){2-4} \cmidrule(r){6-9} 
        &PART C&492& 282/210 &&\textbf{5439}&\textbf{200}&1024$\times$768&\makecell[c]{Hap (803) Dis (1801) Fear (634) \\Ang (466) Con (279) Sur (878) \\Sad (374) Oth (204)}& \\
        \bottomrule
        \end{tabular}
        \begin{tablenotes}
        \footnotesize
        \item[1] Some databases contain not only MEs but also MaEs, as well as long video clips for the detection task. But here we only show the information about ME data. Note that all statistical data are from the corresponding original paper or downloaded databases.
        \item[2] The number of participants was counted based on the data given in the corresponding original paper, but some participants were not successfully induced to make MEs.
        \item[3] Pos: Positive; Neg: Negative; Sur: Surprise; Amu: Amusement; Hap: Happiness; Dis: Disgust; Rep: Repression; Ang: Anger; Sad: Sadness; Con: Contempt; Unc: Unclear; Oth: Others; PosSur: Positively surprise; NegSur: Negatively surprise; RepSur: Repressively surprise; PosRep: Positively repression; NegRep: Negatively repression.
        \end{tablenotes}
        
    \end{threeparttable}
\end{table*}

    \subsection{Micro-expression Databases}
    The premise of obtaining an automatic MER algorithm with excellent performance is to hold a database with sufficient ME samples whose labels are credible and whose visual features are distinguishable. As an emerging field of affective computing, the number of ME databases is still relatively limited. Nevertheless, since more and more researchers have begun to pay attention to ME analysis, some high-quality databases are gradually springing up. Table~\ref{tab_ExistDatabases} clearly summarizes the characteristics of these databases.

    As the two earliest proposed ME databases, samples in the USF-HD~\cite{shreve2011macro} and Polikovsky~\cite{polikovsky2009facial} databases are all posed MEs. The participants were first required to watch video clips containing ME samples and then posed them by imitation. However, naturally generated MEs strongly correlate with emotions, while the posed ones are deliberately displayed and have nothing to do with the current emotional state of the participants. Consequently, these two databases are rarely used by researchers for ME analysis.

    The subsequent researchers proposed to induce spontaneous MEs with the neutralization paradigm. Under this paradigm, several strong emotional stimuli were employed to elicit expressions. Participants were endowed with a certain degree of high-pressure mechanism, and instructed to keep a neutral face as much as possible. Databases adopting the neutralization paradigm include SMIC\cite{li2013spontaneous}, CASME\cite{yan2013casme}, CASME \uppercase\expandafter{\romannumeral2}\cite{yan2014casme}, CAS(ME)$^2$\cite{qu2017cas}, SAMM\cite{davison2016samm}, MMEW\cite{ben2021video}, and 4DME\cite{li20224dme}, which are to be introduced in turn.

    SMIC database\cite{li2013spontaneous} is the first published spontaneous ME database, which consists of three parts: HS, VIS, and NIR. The HS part includes 164 ME samples from 16 participants,  recorded by a high-speed camera with a frame rate of 100 frames per second (fps) and a resolution of 640$\times$480. Both the VIS and NIR parts contain 71 ME samples from 8 individuals, while the former part was recorded using a standard visual camera and the latter using a near-infrared camera. Two annotators classified each ME into three emotion categories (\textit{positive}, \textit{negative}, and \textit{surprise}) based on the participants' self-reports about the elicitation videos. Facial AUs were not annotated in SMIC.

    CASME series databases are released by the Institute of Psychology, Chinese Academy of Sciences. As the earliest database in this series, CASME~\cite{yan2013casme} contains a total of 195 ME samples from 19 participants with a frame rate of 60fps. Two annotators labeled the facial AUs, together with the corresponding onset, apex, and offset frames of each ME sample frame by frame. According to the facial AUs, participants' self-reports, and the relevant video content, MEs were divided into eight emotion categories: \textit{amusement}, \textit{sadness}, \textit{disgust}, \textit{surprise}, \textit{contempt}, \textit{fear}, \textit{repression}, and \textit{tense}. CASME \uppercase\expandafter{\romannumeral2}~\cite{yan2014casme} is an advanced version of CASME. First, the number of ME samples in CASME \uppercase\expandafter{\romannumeral2} has been expanded to 247 samples from 26 participants. Besides, CASME \uppercase\expandafter{\romannumeral2} provides a higher frame rate of 200fps and facial area resolution of 280$\times$340 to capture more subtle changes in expressions. Five emotion categories were labeled in CASME \uppercase\expandafter{\romannumeral2}: \textit{happiness}, \textit{disgust}, \textit{surprise}, \textit{repression}, and \textit{others}. The CAS(ME)$^2$ database~\cite{qu2017cas} embodies two parts, both of which were collected at 30fps and 640$\times$480 pixels. Different from all the other databases above, there are 87 long video clips containing both MaEs and MEs in the first part of CAS(ME)$^2$, which can be used to promote the research of ME detection. The other part consists of 300 MaEs and 57 MEs, which were labeled with four emotion tags, including \textit{positive}, \textit{negative}, \textit{surprise}, and \textit{others}.

    SAMM database~\cite{davison2016samm} has the highest resolution of all published spontaneous ME databases, which includes 159 ME samples generated by 32 participants, with a frame rate of 200fps and a resolution of 2040$\times$1088. Different from other databases, to achieve a better elicitation effect, participants were asked to fill in a scale before the formal start of the collection, and then a series of stimulus videos were customized for each participant according to the scale. SAMM contains seven emotion categories: \textit{happiness}, \textit{disgust}, \textit{surprise}, \textit{fear}, \textit{anger}, \textit{sadness}, and \textit{others}. Three coders annotated the AUs and key-frames in detail for each ME sample.

    MMEW database~\cite{ben2021video} consists of 300 ME and 900 MaE samples from 36 participants, which were collected with 90 fps and 1920$\times$1080 resolution. Each expression sample is marked with seven emotion labels (the same as SAMM), AUs, and three key-frames. Compared with the previous databases, MMEW is more conducive to the models using the MaE samples under the same parameter setting and elicitated environment to assist in learning ME features. 

    To comprehensively capture the movement information of ME in all directions as much as possible, 4DME database~\cite{li20224dme} has made significant innovations in the recording method. Each ME sample in this database has multi-modality video data, including 4D facial data reconstructed by 3D facial meshes sequences, traditional 2D frontal facial grayscale, RGB and depth videos. 4DME contains 267 MEs and 123 MaEs from 41 participants, thus 1,068 ME videos of four forms and 492 MaE videos in total. In addition, five emotion labels (\textit{positive}, \textit{negative}, \textit{surprise}, \textit{repression}, and \textit{others}) were annotated based on facial AUs only, noting that each sample may have multiple emotion labels (up to two).

    Unlike databases with the neutralization paradigm, the MEVIEW database\cite{husak2017spotting} consists of video clips of two real high-pressure scenes downloaded from the Internet. There are 29 ME samples in total, with a frame rate of 30fps and a resolution of 1280$\times$720, divided into seven emotion categories (the same as SAMM) with manual annotation. Although these samples are from actual life scenarios and have high ecological validity, there are many uncontrollable factors, such as frequent camera shot switching, which results in fewer segments containing full human faces.

    The CAS(ME)$^3$ database\cite{li2022cas} adopted the mock crime paradigm to elicit MEs with high ecological validity. However, unlike MEVIEW, the collection was still controlled in the laboratory environment, yielding 166 MEs and 347 MaEs. CAS(ME)$^3$ also contains two other parts: one consists of 943 MEs and 3,143 MaEs collected using the neutralization paradigm, respectively marked with AUs, key-frames, and seven emotion labels (the same as SAMM) for each sample; the other part contains 1,508 unlabeled long video clips, which can be used for the self-supervised learning task of ME detection and recognition. This database was collected at a frame rate of 30fps with a resolution of 1280$\times$720.
    
    Despite more and more databases striving to record the movement characteristics of MEs more detailedly and comprehensively through various methods, these databases are still small-scale databases. In automatic ME analysis, models based on deep learning have become mainstream. However, due to the insufficient sample size, the complexity of the model can easily lead to overfitting in the training process\cite{Kumar2021MicroExpressionCB,takalkar2021lgattnet,pan2021micro}. Though we can alleviate this problem by using data augmentation to increase the number of samples, many uncontrollable noises might be introduced. Some work has proposed using composite databases to train the model\cite{yap2018facial, see2019megc, zong2019cross}, but different databases have different parameter settings, and thus such a simple fusion is not reasonable. In addition, due to the short duration and low intensity of MEs, a higher frame rate may contribute to capturing more details. Nevertheless, the highest frame rate of all above databases is only 200fps, and most are less than 100fps. Therefore, it is necessary to establish a larger-scale ME database with a higher frame rate.
    
    \subsection{Micro-expression Recognition Approaches}
    In the past decade, MER has gained increasing attention from researchers in affective computing and computer vision. The first attempt at automatic, spontaneous MER dates back to 2011, Pfister et al.~\cite{pfister2011recognising} utilized a local binary pattern from three orthogonal planes~(LBP-TOP) to explore MER on the first spontaneous ME database SMIC. Since then, there have been numerous efforts dedicated to developing automatic MER techniques. Generally, current MER methods can be broadly categorized into hand-crafted and deep learning methods. Typical hand-crafted ME features include LBP-TOP~\cite{zhao2007dynamic}, HOOF~\cite{chaudhry2009histograms}, 3DHOG~\cite{polikovsky2009facial}, and their variants~\cite{wang2015lbp, huang2016spontaneous,huang2017discriminative,li2017towards,xu2017microexpression}. However, the hand-crafted methods heavily rely on complex expert knowledge, and the extracted ME features have limited discrimination. Current MER methods mainly employ deep neural networks for high-level expression feature learning and emotion classification, with a focus on addressing the challenges of subtle ME and insufficient ME data for model training. Furthermore, 
     current deep learning MER methods can be divided into single frame-based and video sequence-based approaches based on whether they fully consider the temporal information of ME. In the following sections, we will categorize and summarize these two types of MER methods.
    
    \subsubsection{Single frame-based MER methods.}
    The single frame-based MER method typically utilizes only the highest intensity frame, i.e., the apex frame in terms of RGB or optical-flow format, from the ME video as input for neural networks to learn the ME features. After considering the challenge of lacking sufficient ME samples, Peng et al.~\cite{peng2018macro} first chose ResNet-10~\cite{he2016deep}, which was pre-trained on a large-scale image database, as the backbone, and then continued to fine-tune the classification network on large MaE samples for MER using only apex frames. Encouragingly, the recognition accuracy exceeds the hand-crafted methods based on LBP-TOP, HOOF, and 3DHOG. Inspired by the success of capsule models on image recognition, Quang et al.~\cite{van2019capsulenet} proposed a CapsuleNet for MER using only apex frames. Recently, Xia et al.~\cite{xia2020learning} proposed an expression-identity disentangle network for MER by leveraging MaE databases as guidance. Li et al.~\cite{li2020joint} first spotted the apex frame by estimating pixel-level change rates in the frequency domain, and then proposed a joint feature learning architecture coupling local and global information from the detected apex frames to recognize MEs. 
    
    At the same time, Liong et al.~\cite{liong2018less} explored the effectiveness and superiority of using the optical flow of the apex frame in ME video. Inspired by this work, Liu et al.~\cite{liu2019neural} first calculated the optical-flow image of the apex frame to the onset frame in the ME clips and then used the pre-trained ResNet-18 network to encode the optical-flow image for MER. In particular, they introduced domain adversarial training strategies to address the challenge of lacking large-scale ME data for training and won first place for MEGC2019. Furthermore, Zhou et al.~\cite{zhou2022feature} proposed a novel Feature Refinement (FR) with expression-specific feature learning and fusion for MER based on optical-flow information of apex frames. Gong et al.~\cite{Gong2022MetaMMFNetMB} proposed a meta-learning based multi-model fusion network for MER. Liu et al.~\cite{Liu2022MicroexpressionRB} proposed a novel MER framework with a SqueezeNet~\cite{Iandola2016SqueezeNetAA} for spotting the apex frame and a 3D-CNN for recognition. 
    
     Overall, single frame-based MER investigations are conducted on apex frames of ME videos, which can reduce the complexity of the used deep neural networks. Additionally, this method has the benefit of utilizing large-scale images for transfer learning to effectively address model overfitting due to insufficient ME data. However, it should be noted that single frame-based MER disregards the temporal information present in ME videos, which contains valuable clues and is a crucial feature that distinguishes MEs from MaEs.
     
    \subsubsection{Video sequence-based MER methods.}
    Unlike the single frame-based MER, video sequence-based MER has the ability to learn spatiotemporal ME feature from the entire ME video or its subsequences. As a result, video sequence-based MER is often preferred over single frame-based MER for capturing more detailed information about MEs. After fully considering the significant expression states in the ME video, Kim et al.~\cite{kim2016micro} first used CNN to encode the spatial feature of each expression state~(i.e., onset, onset to apex transition, apex, apex to offset transition and offset), then utilized LSTM to learn the temporal features based on the extracted spatial ME features. Wang et al.~\cite{wang2018micro} proposed Transferring Long-term Convolutional Nerual Network~(TLCNN) to solve the learning of spatial-temporal ME feature under small sample ME data. The TLCNN was also based on the CNN-LSTM structure and transferred knowledge from large-scale expression data and single frames of ME video clips. Khor et al.~\cite{khor2018enriched} proposed an Enriched Long-term Recurrent Convolutional Network~(ELRCN) which made spatial and temporal enrichment by stacking different input data and features. Unlike the CNN-LSTM architecture, 3D convolution neural network~(3D-CNN)~\cite{ji20123d} can simultaneously learn the spatial and temporal ME features. Based on 3D-CNN, Peng et al.~\cite{peng2017dual} proposed a Dual Temporal Scale Convolutional Neural Network~(DTSCNN), which used the optical-flow sequences of ME videos as model input to obtain high-level ME features and can adapt to a different frame rate of ME video clips. Wang et al.~\cite{wang2020eulerian} proposed an MER framework based on Eulerian motion based 3D-CNN~(EM-CED), which used the pre-extracted Eulerian motion feature maps as input and with a global attention module to encode rich spatiotemporal information. Xia et al.~\cite{xia2019spatiotemporal} proposed a deep recurrent convolutional networks based MER approach, which modeled the spatiotemporal ME deformations in views of facial appearance and geometry separately. To solve the challenge of extracting high-level ME features from the training model lacking sufficient and class-balanced ME samples, Zhao et al.~\cite{zhao2021two} extracted the ME optical-flow sequence to express the original ME video and proposed a novel two-stage learning (i.e., prior learning and target learning) method based on a siamese 3D-CNN for MER. Sun et al.~\cite{sun2020dynamic} proposed a knowledge transfer technique that distilled and transferred knowledge from AUs for MER based on crucial temporal sequences, where knowledge from a pre-trained deep teacher neural network was distilled and transferred to a shallow student neural network. Zhao et al.~\cite{Zhao2022MEPLANAD} proposed a deep prototypical learning framework on RGB key-frame sequences, namely ME-PLAN, based on a 3D residual prototypical network and a local-wise attention module for MER. Recently, with the advancement of deep learning technology, some excellent neural networks, such as GCN~\cite{Xie2020AUassistedGA, Lei2021MicroexpressionRB, Kumar2021MicroExpressionCB, 10219571} and Transformers ~\cite{Hong2022LateFV}, ~\cite{nguyen2023micron}, have also been used for MER.

    Although video sequence-based MER leverages the spatial-temporal information of ME, the corresponding models tend to have higher structural complexity and are prone to overfit on current small-scale ME databases~\cite{zhao2021two,Zhao2023FacialMA,Li2021DeepLF}. Consequently, building a large-scale ME database remains a crucial task in developing an automatic MER system, as it serves as a fundamental component.
    
\begin{figure*}[htbp]
    \centering
    \includegraphics[scale=0.65]{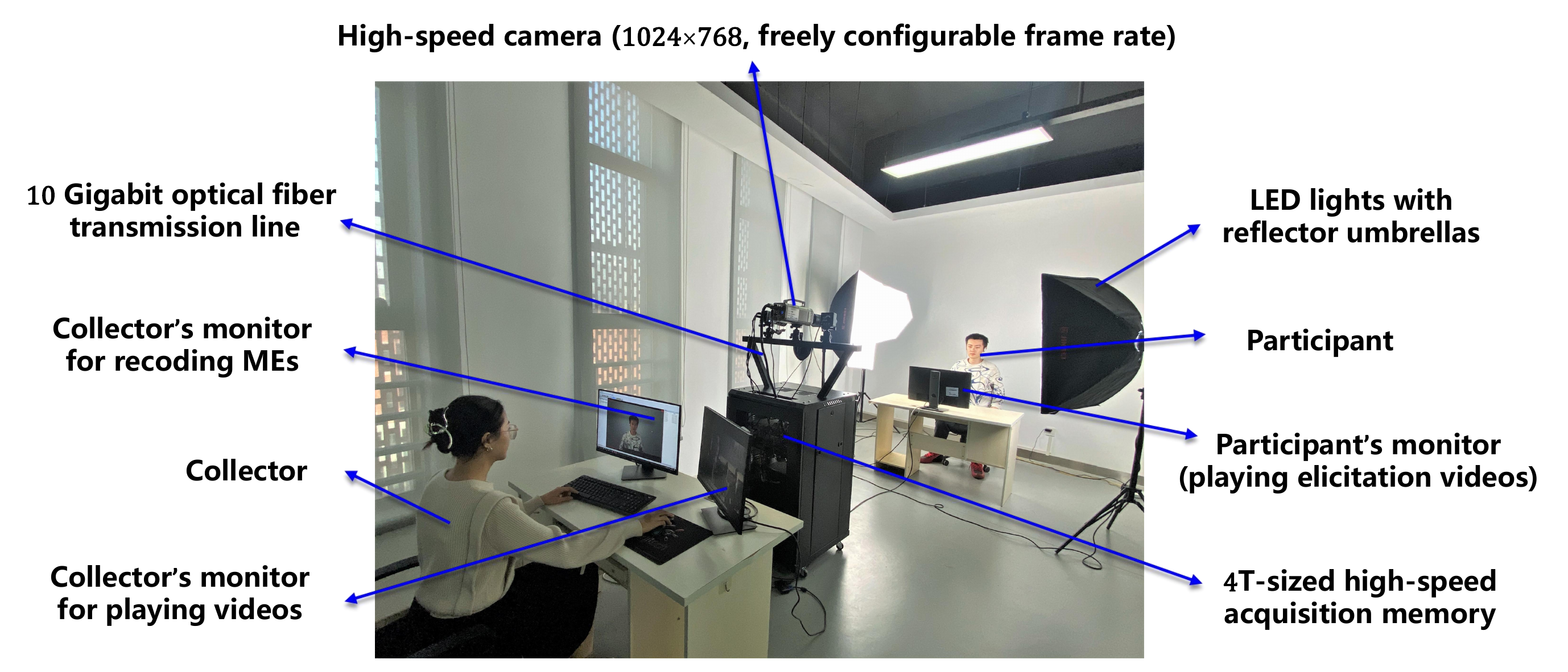}
    \caption{Experimental environment for eliciting MEs}
    \label{env}
\end{figure*}
  
\section{DFME Database Profile}

    As the old saying goes, 'One cannot make bricks without straw'. To address the problem of ME data hunger, we construct a database of spontaneous ME with the largest sample size at present, called DFME. In the following subsections, we will elaborate on the building details and statistical properties of our DFME database.
    
    \subsection{Participant and Equipment}
    In our DFME, 671 participants were recruited (381 males and 290 females), mainly for college students and teaching staff. Participants were age-distributed between 17 and 40 years, with a mean age of 22.43 years (standard deviation = 2.54), and all from China. Before starting the formal experiment, the participants were informed about the purpose, experimental procedure, possible benefits and risks of our research. All studies involving human participants (which are all ordinary people not involving patients) adhered to the Declaration of Helsinki. 
    Everybody participating in the experiment signed informed consent and chose whether to allow their facial images and videos used for the academic paper, ensuring ethical and responsible research practices.
    
    Given the subtle nature and brief duration of MEs, the recording process is susceptible to disturbances from external factors. Therefore, we conducted the recording in a well-controlled laboratory environment, as depicted in Fig.~\ref{env}. Three LED lights equipped with reflector umbrellas were strategically positioned to ensure a consistently bright and stable light source illuminating the participants' faces during experiments. In addition, we employed a self-developed high-speed camera (1024$\times$768, freely configurable frame rates) for capturing MEs, which was connected via a 10 Gigabit optical fiber transmission line to a 4T-sized high-speed acquisition memory, facilitating real-time storage of the collected ME video clips.
    
    \begin{table}[t]
    \caption{Video clips for eliciting MEs}\label{tab_Elima}
    \centering
        \begin{tabular}{cccc} 
        \toprule
        Video ID & During Time & Emotion  Category & Mean Score(1-5)\\\midrule
        02sa & 3'44'' & Sadness & 4\\
        03sa & 4'18'' & Sadness & 3.36\\
        06c & 2'01'' & Contempt & 2.83\\
        07a & 1'26'' & Anger & 3.49\\
        08su & 1'26'' & Surprise & 2.16\\
        09f & 2'22'' & Fear & 3.72\\
        10a & 2'58'' & Anger & 4.33\\
        11d & 1'24'' & Disgust & 3.95\\
        13f & 2'14'' & Fear & 3.36\\
        14d & 1'22'' & Disgust & 3.23\\
        17h & 1'17'' & Happiness & 2.81\\
        18h & 1'58'' & Happiness & 3.08\\
        20d & 0'46'' & Disgust & 2.87\\
        21c & 1'44'' & Contempt & 2.11\\
        23sa & 1'44'' & Sadness & 3.25\\
        \bottomrule
        \end{tabular}
    \end{table}
    \subsection{Elicitation Process}
    So far, there are three generations of ME-eliciting paradigms as outlined in~\cite{li2022cas}. Although the third generation has the highest ecological validity, it is inevitable to interact and have conversations with the participants when simulating the natural scenes. These irrelevant body and mouth movements caused by speaking are also a kind of noise for MEs. Hence, we still employ the neutralization paradigm for ME elicitation to minimize noise interference and focus more on the movement characteristics of MEs. 
    Specific details of the elicitation process will be introduced below.
    
    \subsubsection{Elicitation Materials}
    The effectiveness of elicitation materials determines the quantity and quality of MEs, so selecting the materials with high emotional valence is very crucial~\cite{yan2014casme}. The stimuli we used were all video clips from the Internet, ranging in length from 46 seconds to 258 seconds. In order to find more effective stimulus materials, we recruited 50 volunteers to evaluate 30 video clips collected previously. The evaluation process was as follows: after watching each video, volunteers were asked to choose only one emotion from \textit{happiness, anger, contempt, disgust, fear, sadness} and \textit{surprise} as the main emotion evoked by this video, and score the stimulus level on an integer scale of 1 to 5, corresponding to the intensity from mildest (but not None) to strongest. Finally, we took the emotion selected by more than half of the volunteers as the emotional class of each video. By ranking the average stimulus intensity values, we obtained the optimal 15 video clips as elicitation materials adopted in our experiment. Specific statistical details are shown in Table~\ref{tab_Elima}.
        
    \subsubsection{Elicitation and Collection Procedure}
    The collection took place in a meticulously arranged laboratory setting. Prior to start, each participant was seated at an assigned location. Through adjustments to the seat's height, the camera's focal length and the LED lamps' brightness, we ensured that the participant’s face appeared utterly, clearly, and brightly in the centre of the screen.
    
    The monitor in front of the participant would play ten randomly selected elicitation videos~(EVs) covering all seven discrete emotional types that had been previously verified effective in turn. The collector synchronously managed the recording of the participant's facial region through the high-speed camera to capture facial videos~(FVs) containing ME fragments. While watching EVs, participants were required to maintain a neutral face as far as possible and control the occurrence of their facial expressions, alongside keeping an upright sitting posture, avoiding excessive head movements, and dedicating complete attention to the played EV. If they failed and repeatedly showed obvious expressions, they would have to complete a lengthy and tedious questionnaire as punishment. 
    
    After watching each EV, participants would have a period of rest to ease their emotions. Meanwhile, they were instructed to fill in an affective grade scale according to the emotional experience generated just now, and form a self-report detailing the timestamp of any observed expressions, emotion category and intensity. Specifically, the collector replayed from the beginning of the participant's FV latest collected and confirmed the timestamp $t$ in the corresponding EV which coincided with the appearance of a facial movement segment. Once $t$ was determined, the replay of FV was paused. Participants were inquired about their genuine psychological responses related to the EV within a 3-second time window centered around $t$, and noted down the timestamp $t$ together with their corresponding emotions on the designated areas of the self-report scale. Then, the collector continued to replay the FV and repeated aforementioned steps, until the entire FV playback concluded, so that participants were allowed to view the next EV.
    
    Due to the existence of cognitive differences, the emotional orientation of the elicitation materials and the internal emotional experience of participants are sometimes not exactly consistent. What’s more, external expressions of the same emotion are also diverse on account of individual differences. Therefore, it is worth noting that we specifically required participants to clarify their true internal emotions in their self-reports whenever facial expressions appear, which is vital in aiding subsequent annotators to comprehend the nuances of their MEs.
    
\begin{table*}[t]
    \caption{Key AUs Included in DFME}\label{tab_AU}
    \centering
        \begin{tabular}{cccccccc} 
        \toprule
        \multicolumn{2}{c}{Upper Face Action Units}&\multicolumn{4}{c}{Lower Face Action Units}&\multicolumn{2}{c}{Miscellaneous Actions}\\
        \cmidrule(r){1-2} \cmidrule(r){3-6} \cmidrule(r){7-8}
        AU1 & Inner Brow Raiser & AU9 & Nose Wrinkler & AU18 & Lip Pucker & AU31 & Jaw Clencher\\
        AU2 & Outer Brow Raiser & AU10 & Upper Lip Raiser & AU20 & Lip Stretcher & AU38 & Nostril Dilator\\
        AU4 & Brow Lowerer & AU12 & Lip Corner Puller & AU23 & Lip Tightener & AU39 & Nostril Compressor\\
        AU5 & Upper Lid Raiser & AU14 & Dimpler & AU24 & Lip Presser & M57 & Head Forward\\
        AU6 & Cheek Raiser & AU15 & Lip Corner Depressor & AU25 & Lips Part & M58 & Head Back\\
        AU7 & Lid Tightener & AU16 & Lower Lip Depressor & AU28 & Lip Suck & & \\
         & & AU17 & Chin Raiser & & & & \\
        \bottomrule
        \end{tabular}
\end{table*}
                
    \begin{figure*}[h]
        \centering
        \includegraphics[scale=0.57]{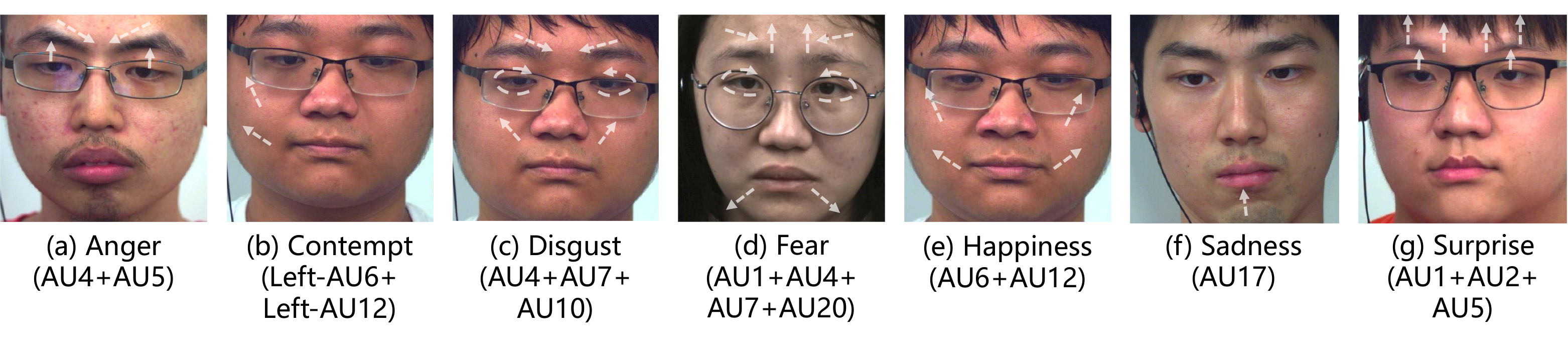}
        \caption{Representative ME Samples of Seven Discrete Emotion Categories in DFME}
        \label{exam}
    \end{figure*}

    \subsection{ME Annotation}
    Building the DFME database required a two-stage annotation: the sample selection stage as well as the coding and category labeling stage. In the first stage, we clipped short fragments containing valid expression samples from the collected long video sequences. The second stage included three successive rounds of fine-grained annotation, through which we confirmed all MEs and labeled their three key-frames~(i.e., onset, apex, and offset frame), facial AUs, and emotion categories. Furthermore, we performed annotation agreement test to verify the reliability of emotion labels.
    
    \subsubsection{Sample Selection}
    In the sample selection stage, participants' FV sequences were manually segmented into several shorter video fragments, each capturing at least one ME or MaE.Using the self-developed video annotation software, an experienced annotator checked through the collected original FVs frame by frame to locate the fragments of facial muscle movements. 
    With the guidance of the self-reports from participants, the annotator was able to effectively distinguish the facial expressions definitely related to emotion, and abandon interference data unrelated to emotion (such as violent blinking caused by dry eyelids, habitual mouth opening, etc.). Besides, we reserved some fragments with blinking or eye movements if they contained MaE or ME data.

    \subsubsection{Coding and Category Labeling}
    The apex frame corresponds to the moment when facial expression changes most dramatically. In the first round of the fine-grained annotation, five annotators independently marked out the onset, apex, and offset frame of each expression clip, and the median value of their annotation results was determined as the final result of the three key-frames. Then we filtered the expressions whose duration from onset to offset frame was less than 500ms or from onset to apex frame was less than 250ms as the ME samples, and those out of the time limit were considered as the samples of MaEs. For instance, MEs collected at a frame rate of 500fps should meet either $f_{offset}-f_{onset}+1\leq250$ or $f_{apex}-f_{onset}+1\leq125$, where $f_k$ represents the moment index corresponding to the key-frame $k$.
    
    In the second round of fine-grained annotation, we mainly annotated the AUs that occurred in MEs using the Facial Action Coding System (FACS)\cite{ekman1978facial}. There may exist a single AU such as AU4, or a combination of more different AUs like AU6+AU12 in an ME. When multiple categories of AUs appear, some obscure ones are easily overlooked. To enhance the reliability and integrity of the AU labels, two experienced FACS-certified annotators independently labeled the AUs for all the MEs identified previously. According to the actual induction of the participants during the experiments, and also referring to the AUs mainly involved in the previously published ME databases, we totally included 24 different categories of AUs for annotation. Of these AUs, six categories appear in the upper face, 13 in the lower face, and the other five belong to miscellaneous actions. Table \ref{tab_AU} lists the specific AU numbers and their corresponding face actions. Since the manually annotated AU intensity is highly subjective, annotators merely indicated whether each AU appeared during the annotation rather than defining the intensity of its occurrence.
    
    After labeling the AUs, the two annotators determined the final AU label through crosscheck and discussion. The reliability $R$ between the two annotators was 0.83, which was calculated as
    \begin{equation}
        R=2\times \frac {AU(A_{1})\cap AU(A_{2})}{All_{AU}}
    \end{equation}
    where $AU(A_{1})\cap AU(A_{2})$ means the number of AUs both annotators agreed, and $All_{AU}$ is the total number of AUs in an ME labeled out by the two annotators.
    
    In the third round of fine-grained labeling, we performed the emotion labeling of MEs taking eight categories into account: \textit{happiness}, \textit{anger}, \textit{contempt}, \textit{disgust}, \textit{fear}, \textit{sadness}, \textit{surprise}, and \textit{others}. '\textit{Others}' represents MEs that are difficult to divide into the former seven prototypical emotion categories. Seven annotators independently gave the emotion labels of all MEs. When disagreements arised, a '50\% majority voting' approach was employed, where a sample was assigned a specific emotion label if at least four annotators agreed on that label. For samples with unresolved disagreements, a second round of voting was conducted through collective discussions among all annotators to determine the emotion label. If a consensus still cannot be reached, the sample was categorized as '\textit{Others}'.
    
    In previous spontaneous ME databases, the reference basis of emotion labeling was not precisely the same. In some databases, as represented by SMIC, emotion labels were determined based on self-reports provided by participants. Some other studies believed that seeing is believing, so their annotation was based on the correspondence between AUs and emotions. However, on the one hand, unlike MaEs, only part of the AUs can appear simultaneously in MEs due to their low intensity, and some AUs are shared by different emotion categories, which may lead to category confusion. On the other hand, we should not ignore the differences in self-emotional cognition of different participants, which means that the self-reports given for the whole piece of elicitation materials may be rough and inaccurate. Therefore, in DFME, the emotion labels were determined through a comprehensive analysis of facial AUs, self-reports of participants, and elicitation material contents, which is consistent with the method adopted by the CASME series. It is worth mentioning that we obtained the participants' fine-grained self-reports in the data collection process, and this is also the information that we recommend as a priority for reference when determining emotion labels. We matched the corresponding timestamps of MEs and elicitation materials through playback, enabling participants to report their emotions for each time of successful ME induction, which significantly improved the confidence of self-reports in emotion labeling. Fig. \ref{exam} shows some representative ME samples of seven discrete emotion categories in DFME.
   
   \subsubsection{Annotation Agreement}
    
    Having reliable emotion categories of MEs is of vital significance for a database. In this section, we utilized Fleiss's Kappa test\cite{fleiss1971measuring} to evaluate the quality of our emotion annotation encouraged by work\cite{jiang2020dfew}. Fleiss's Kappa is a measure of the agreement among three or more annotators, testing the consistency of annotation results. Therefore, we consider Fleiss's Kappa as an excellent indicator to evaluate the reliability of emotion annotation.

    In DFME, seven annotators independently labeled each ME sample based on facial AUs, an accurate self-report, and the corresponding elicitation material content. The samples were divided into eight emotion categories: \{1: happiness, 2: anger, 3: contempt, 4: disgust, 5: fear, 6: sadness, 7: surprise, 8: others\}. Let $n=7$ represent the total number of annotation personnel, $N$ indicate the total number of ME video clips, $K=8$ represent the number of emotion categories. $n_{ij}$ is the number of annotators who assigned the $i$-th ME video clip to the $j$-th category, so we can calculate $p_{j}$, the proportion of all assignments which were to the $j$-th emotion:
    \begin{equation}
        p_{j}=\frac{1}{N \times n} \sum_{i=1}^{N} n_{ij},
    \end{equation}
    \begin{equation}
        \sum_{j=1}^{K} p_{j}=1.
    \end{equation}

    Then, the extent of agreement among the $n$ annotators for the $i$-th ME video clip indicated by $P_{i}$ is calculated. In other words, it can be indexed by the proportion of pairs agreeing in their evaluation of the $i$-th ME out of all the $n(n-1)$ possible pairs of agreement:
    \begin{equation}
        P_{i}=\frac{1}{n \times (n-1)} [(\sum_{j=1}^{K} n_{ij}^{2}) - n].
    \end{equation}

    The mean of $P_{i}$ is therefore:
    \begin{equation}
        \overline{P}=\frac{1}{N} \sum_{i=1}^{N} P_{i}.
    \end{equation}

    And we also have $\overline{P_{e}}$:
    \begin{equation}
        \overline{P_{e}}=\sum_{j=1}^{K} p_{j}^{2}.
    \end{equation}

    Finally, we can calculate $\kappa$ by:
    \begin{equation}
        \kappa = \frac{\overline{P}-\overline{P_{e}}}{1-\overline{P_{e}}}.
    \end{equation}
           
    \begin{table}[htbp]
    \caption{Interpretation of $\kappa$ for Fleiss'Kappa Test}\label{tab1}
    \centering
        \begin{tabular}{cc} 
        \toprule
        $\kappa$&Interpretation\\\midrule
        $\leq$ 0 & Poor agreement\\
        0.01-0.20 & Slight agreement\\
        0.21-0.40 & Fair agreement\\
        0.41-0.60 & Moderate agreement\\
        0.61-0.80 & Substantial agreement\\
        0.81-1.00 & Almost perfect agreement\\
        \bottomrule
        \end{tabular}
    \end{table}
    
    Thus, we obtained $\kappa=0.72$ through performing Fleiss’s Kappa test in DFME. According to Table \ref{tab1}, we know that all of our emotion annotators achieve substantial agreement, meaning that our emotion labels are quite reliable.
   
    \subsection{Statistical Properties of DFME}
        
    \begin{table*}[t]
    \caption{Distribution of AU Labels in DFME}\label{AU_count}
    \centering
        \begin{threeparttable}
        \setlength{\tabcolsep}{2mm}
        \begin{tabular}{cccccccccccc} 
        \toprule
        AU & number & AU & number & AU & number & AU & number & AU & number & AU & number\\
        \cmidrule(r){1-2} \cmidrule(r){3-4} \cmidrule(r){5-6} \cmidrule(r){7-8} \cmidrule(r){9-10} \cmidrule(r){11-12}
        AU1 & 876 & AU9 & 49 & AU17 & 408 & AU28 & 18 & L/R-AU1\tnote{1} & 280 & L/R-AU10 & 154\\
        AU2 & 708 & AU10 & 283 & AU18 & 24 & AU31 & 28 & L/R-AU2 & 367 & L/R-AU12 & 460\\
        AU4 & 2617 & AU12 & 712 & AU20 & 77 & AU38 & 173 & L/R-AU4 & 190 & L/R-AU14 & 50\\
        AU5 & 973 & AU14 & 470 & AU23 & 235 & AU39 & 35 & L/R-AU5 & 156 & L/R-AU15 & 31\\
        AU6 & 649 & AU15 & 116 & AU24 & 757 & M57 & 4 & L/R-AU6 & 93 & L/R-AU20 & 17\\
        AU7 & 1624 & AU16 & 35 & AU25 & 60 & M58 & 26 & L/R-AU7 & 173 & Total & 12928\\
        \bottomrule
        \end{tabular}
        \begin{tablenotes}
        \footnotesize
        \item[1] L/R means the Left/Right half part of an AU.
        \end{tablenotes}
        \end{threeparttable}
    \end{table*}
         
     \begin{figure*}[h]
        \centering
        \includegraphics[scale=0.75]{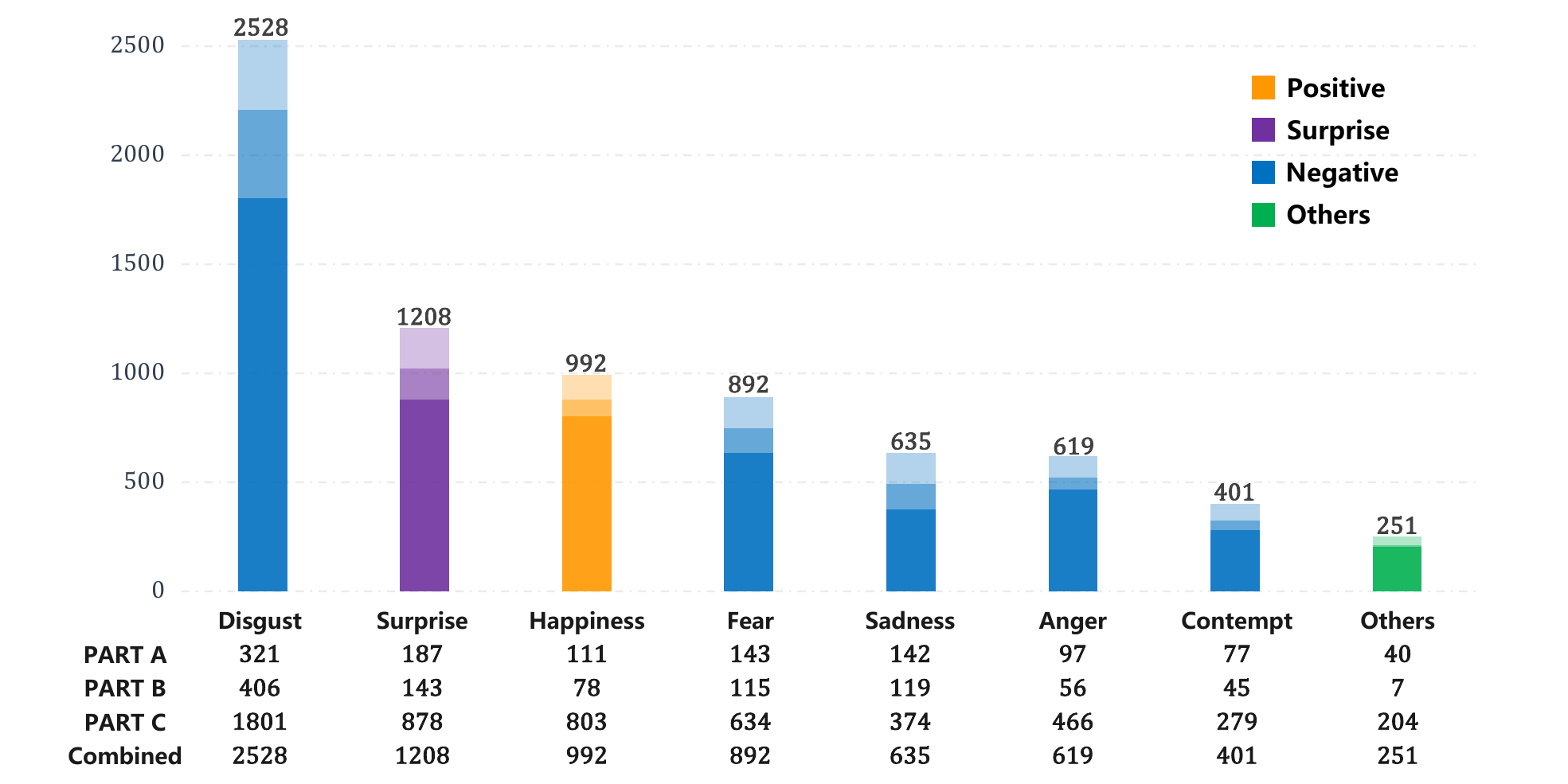}
        \caption{Distribution of ME Samples in DFME. Each column represents the total sample number of an emotion category, and the three pieces colored from light to deep show the proportion of samples in PART A, PART B, and PART C, respectively.}
        \label{stat}
    \end{figure*}
 
    \begin{table*}[htbp]
    \caption{AUs of High Occurrence in MEs of Seven Discrete Emotion Categories}\label{AU_percent}
    \centering
        \begin{threeparttable}
        \setlength{\tabcolsep}{2mm}
        \begin{tabular}{cccccccccccccc} 
        \toprule
        \multicolumn{2}{c}{Happiness}&\multicolumn{2}{c}{Anger}&\multicolumn{2}{c}{Contempt}&\multicolumn{2}{c}{Disgust}&\multicolumn{2}{c}{Fear}&\multicolumn{2}{c}{Sadness}&\multicolumn{2}{c}{Surprise}\\
        \cmidrule(r){1-2} \cmidrule(r){3-4} \cmidrule(r){5-6} \cmidrule(r){7-8} \cmidrule(r){9-10} \cmidrule(r){11-12} \cmidrule(r){13-14}
        AU & pct(\%)\tnote{1} & AU & pct(\%) & AU & pct(\%) & AU & pct(\%) & AU & pct(\%) & AU & pct(\%) & AU & pct(\%)\\\midrule
        AU12 & 79.8 & AU4 & 72.5 & L/R-AU12 & 78.7 & AU4 & 73.6 & AU4 & 54.1 & AU4 & 42.2 & AU1 & 65.6\\
        AU6 & 61.6 & AU7 & 29.1 & AU6 & 19.2 & AU7 & 40.4 & AU7 & 35.3 & AU14 & 26.1 & AU5 & 60.2\\
        AU24 & 12.1 & AU24 & 16.3 & L/R-AU10 & 10.6 & AU10 & 11.8 & AU5 & 16.2 & AU24 & 19.2 & AU2 & 60.0\\
        L/R-AU12 & 10.1 & AU5 & 7.6 & AU7 & 7.8 & AU24 & 8.4 & AU24 & 14.5 & AU7 & 16.5 & L/R-AU2 & 25.6\\
        AU10 & 6.2 & AU23 & 5.6 & L/R-AU2 & 5.7 & AU14 & 6.7 & AU1 & 11.1 & AU17 & 10.8 & L/R-AU1 & 17.8\\
         & & AU14 & 5.6 & AU14 & 5.7 & & & AU14 & 8.8 & AU15 & 6.9 & L/R-AU5 & 10.7\\
         & & AU10 & 5.2 & & & & & AU17 & 6.0 & AU23 & 5.1 & & \\
         & & AU17 & 4.8 & & & & & AU10 & 4.8 & AU1 & 4.8 & & \\
        \bottomrule
        \end{tabular}
        \begin{tablenotes}
        \footnotesize
        \item[1] percentage(pct): the statistical range is all MEs from the first 300 participants.
        \end{tablenotes}
        \end{threeparttable}
    \end{table*}
 
    The DFME database consists of three parts: PART A, PART B, and PART C. The only difference between these three parts is the frame rate setting of the high-speed camera in the experiment. In PART A, all 1,118 ME samples from 72 participants have a frame rate of 500fps. The frame rate of PART B is 300fps with 969 ME samples from 92 participants. PART C has the most data size with 5,439 ME samples from 492 participants, whose frame rate is 200fps. Although we recruited a total of 671 participants, 15 of them had strong control over their facial expressions, from whom we could not collect any ME sample. Therefore, the final DFME database contains 7,526 ME samples from 656 participants, and we gave each sample an emotion category label as well as AU labels annotated according to FACS. Fig. \ref{stat} describes the distribution of ME samples in detail. According to the specific distribution of AU labels shown in Table \ref{AU_count}, the average number of AU's per ME sample occurrence can be calculated as $12928/7526 = 1.718$. 
    
    Given that we have collected the fine-grained self-reports and the AU labels with considerable reliability, this fosters the exploration of the emotion-AU correspondence rule in MEs. Therefore, we counted the ratio of high-occurrence AUs in each emotion (Table \ref{AU_percent}), which reflects the existence preference of AU in MEs with different emotions, not affected by the emotional category imbalance problem in the database. We also matched the emotion and AU combinations according to the statistical results, and the conclusions are shown as Table \ref{AUcombine}.
        
    \begin{table}[h]
    \caption{Matching Emotion and AU Combinations in MEs}\label{AUcombine}
    \centering
        \begin{threeparttable}
        \begin{tabular}{cc} 
        \toprule
        Emotion Categories&AU Combinations\\\midrule
        Happiness & AU6+AU12, AU12\\
        Anger & AU4+AU5, AU23\\
        Contempt & L/R-AU12, AU6+L/R-AU12\\
        Disgust & AU4+AU7+AU10, AU14\\
        Fear & AU14+AU24, AU1+AU4, AU4+AU5\\
        Sadness & AU14, AU17, AU15, AU14+AU24\\
        Surprise & AU1+AU2+AU5, AU1+AU2, AU5\\
        Shared\tnote{1} & AU4, AU4+AU7, AU7, AU24\\
        \bottomrule
        \end{tabular}
        \begin{tablenotes}
        \footnotesize
        \item[1] Shared: the AU combinations commonly appearing in Anger, Disgust, Fear and Sadness with high frequency.
        \end{tablenotes}
        \end{threeparttable}
    \end{table}

    Based on the statistical results presented in Table \ref{AU_percent}, we have some findings to discuss:
    \begin{itemize}
      \item In MaEs, AU9 (nose wrinkler) is highly associated with \textit{disgust}, and AU20 (lip stretcher) is related to \textit{fear}. These two AUs frequently appear in MaEs but are not easily induced in MEs. We ought not to conclude that these AUs' association with their corresponding emotions no longer exists in MEs. Instead, when participants tried to restrain their emotions, it was easier to control the movement of certain facial muscles such as AU9 and AU20 than others.
      \item AU4 (brow lowerer), AU7 (lid tightener), and AU24 (lip presser) simultaneously occur at high frequency in different negative emotions (\textit{disgust}, \textit{anger}, \textit{fear}, \textit{sadness}, etc.). Without the assistance of participants' fine-grained self-reports, it is definitely challenging to distinguish MEs of negative emotions merely relying on these common AUs, which is also one of the reasons why some models excessively confuse the \textit{disgust} MEs with those of other negative emotions in the seven-class classification automatic MER task.
      \item In the positive emotion (\textit{i.e., happiness}), some AUs related to negative emotions can occur together with AU6 or AU12, specifically, including AU10 (associated with \textit{disgust}), AU24 (associated with negative emotions), and Left/Right-AU12 (associated with \textit{contempt}). The appearance of these extra AUs is a sign of participants trying to suppress their positive feelings, hide their smiles and twist their expressions.
    \end{itemize}

\section{Database Evaluation}
    In this section, we conducted comprehensive experiments to verify the effectiveness of our DFME database for automatic MER and AU classification tasks, leveraging influential spatiotemporal feature learning and MER models. More specifically, MER involves assigning an emotion class label to a given ME video sample, which is a multi-classification task. On the other hand, AU classification aims to predict whether an AU exists in a video clip, corresponding to a single binary multi-label problem~\cite{varanka2023data}. These experiments can serve as a valuable reference for future research on ME analysis using the DFME database. 
    
    \subsection{Evaluation Database}
    The DFME database is described in detail in Section 3. For the subsequent MER and AU classification verification, we combined $7,275$ samples with clear emotion and AU labels in PART A, B and C of DFME as our experimental database. The emotion labels include \textit{happiness, anger, contempt, disgust, fear, sadness} and \textit{surprise}. By drawing inspiration from CD6ME\cite{varanka2023data}, we also selected the following 12 frequently occurring AUs: \textit{AU1}, \textit{AU2}, \textit{AU4}, \textit{AU5}, \textit{AU6}, \textit{AU7}, \textit{AU9}, \textit{AU10}, \textit{AU12}, \textit{AU14}, \textit{AU15}, and \textit{AU17} as the AU labels for our experiment. In fact, these 12 AUs account for about 88.48\% (11439/12928) of the DFME database, covering most AUs in MEs. Among the remaining AUs, most AUs are small in number, such as AU16 and AU18, while others like AU23 (Lip Tightener) are usually unconscious actions unrelated to emotions. For these reasons, we chose these 12 AUs for the experiments.
    
    \subsection{Data Preprocessing}
    In facial expression recognition, many variables, such as backgrounds and head poses, can affect the final recognition results. Therefore, before formally conducting automatic MER experiments, we need to preprocess all ME videos in the following steps (i.e., face alignment and face cropping) to minimize the influence of irrelevant variables.
    
    \subsubsection{Face Alignment}
    To eliminate variations in pose and angle among all ME samples, we need to perform face alignment. In this step, we took the following operations for each ME sample. Firstly, we selected a frontal face image as a reference and applied Style Aggregated Network (SAN) \cite{SAN} to extract its facial landmarks. Subsequently, we employed Procrustes analysis \cite{GPA} to compute an affine transformation based on the landmarks of the onset frame and those of the reference image. The rationale behind not using landmarks from all frames in the ME video is to avoid errors introduced by the calculation of landmarks and transformations that could significantly impact real MEs. Finally, the transformation was applied to each frame to align the faces. Additionally, some landmarks were situated in regions where MEs may appear, which may not be stable enough for alignment. Therefore, we excluded such landmarks during the alignment process.
    
    \subsubsection{Face Cropping}
    Since the movement of MEs is mainly in the facial area, face cropping is essential to eliminate biases caused by varying backgrounds.
    Following face alignment, we employed RetinaFace\cite{RTF} to crop the faces. Similar to face alignment, face cropping was based on the onset frame rather than each frame of a sample.
 
    \subsection{Evaluation Protocols and Metrics}
    Due to the small sample size of previous databases such as CASME \uppercase\expandafter{\romannumeral2}\cite{yan2014casme}, SAMM\cite{davison2016samm}, and SMIC\cite{li2013spontaneous}, most MER studies employed the leave-one-subject-out strategy for evaluation. However, given the relatively large number of ME clips in DFME, this paper utilized a simpler and more efficient subject-independent 10-fold cross-validation strategy. In each fold, data from 10\% of subjects were selected as the test set, while the remaining 90\% were used for training.
    
    Inline with CD6ME\cite{varanka2023data}, we employed the F1-score as the evaluation metric for AU classification. Furthermore, three widely recognized ME classification metrics, namely Accuracy~(ACC), Unweighted F1-Score~(UF1), and Unweighted Average Recall~(UAR), were utilized to assess the MER performance. Finally, we computed the average of ten folds' outcomes as the final result in the MER.
    
    \subsubsection{Accuracy~(ACC)}
    ACC is one of the most common metrics, which can evaluate the overall performance of the recognition method on the database. It is calculated as follows:
    \begin{equation}
	ACC=\frac{\sum_{i=1}^K TP_i}{\sum_{i=1}^K N_i},
	\end{equation}\label{acc}%
    where $K$ represents the number of the classes, $N_i$ stands for the sample number of the $i$-th class and $TP_i$ is the number of true positive samples of the $i$-th class. 

    \subsubsection{F1-score~(F1)}
    When the problem of class imbalance in the database is pronounced, ACC may not accurately reflect the model's true performance. As a result, F1-score is often employed as the evaluation metric in most classification tasks to address this challenge. The F1-score is defined as shown below:
	\begin{equation}
	F1=\frac{2\cdot TP}{2\cdot TP + FP + FN}.
	\end{equation}\label{F1}%
    where $TP$, $FP$, $FN$ refer to true positives, false positives and false negatives, respectively. Notably, $TP$, $FP$, $FN$ are calculated based on all ten folds in the AU classification task.
    
    \subsubsection{Unweighted F1-score~(UF1)}
    UF1, also known as macro-averaged F1-score, is defined as:
    \begin{equation}
	UF1=\frac{1}{K}\sum_{i=1}^K{\frac{2\cdot TP_i}{2\cdot TP_i + FP_i + FN_i}}.
	\end{equation}\label{UF1}%

    Class imbalance is an intractable problem in the MER task, so introducing UF1 as an evaluation metric can better measure the method's performance in all classes rather than in some major classes.
    
    \subsubsection{Unweighted Average Recall ~(UAR)}
    UAR is also a more suitable metric than ACC when dealing with class imbalance.
    
    \begin{equation}
	UAR=\frac{1}{K}\sum_{i=1}^K\frac{TP_i}{N_i}.
	\end{equation}\label{UAR}
 
    Both UF1 and UAR can effectively assess whether MER methods provide accurate predictions across all classes.

    \subsection{Evaluation Baseline Models}
    To comprehensively validate our database, we specifically selected three different groups of baseline methods for MER and AU classification, including: \textit{3D-CNN Methods, Hand-crafted MER Methods} and \textit{Deep learning MER Methods}.

    \subsubsection{3D-CNN Methods}
    In recent years, many influential 3D-CNN methods have emerged in the field of video classification. Due to their excellent ability to represent spatiotemporal features, they are often used in dynamic ME analysis. Here, we selected 3D-ResNet~(R3D) \cite{Hara2018R3D} and Inflated 3D ConvNet~(I3D)\cite{Carreira2017I3D} as two baseline methods. Hara et al. proposed R3D for tasks such as video classification and recognition. Since then, R3D is often used as the backbone in approaches to video-related tasks. The basic idea of this model is to replace the 2D convolutional kernels with spatiotemporal 3D kernels according to the 2D-ResNet\cite{he2016deep} network structure. I3D\cite{Carreira2017I3D} is based on 2D ConvNet inflation. It utilizes convolutional kernels with different sizes to extract features, and the key idea behind the I3D model is to inflate the 2D filters and pooling kernels into 3D, allowing it to understand and process the spatial and temporal information of video data. 
    
    \subsubsection{Hand-crafted MER Methods}
    The hand-crafted methods in MER are typically based on traditional machine learning, extracting elaborately designed manual features from ME videos. Hand-crafted methods were popular and achieved SOTA results on small ME databases in the early days. We selected LBP-TOP\cite{zhao2007dynamic} and MDMO\cite{liu2015main} as two baseline methods. LBP-TOP extends LBP from 2D to 3D, which extracts features from three orthogonal planes (X-Y, X-T, and Y-T) and connects them together. MDMO refers to the main directional mean optical flow features, which is a typical optical flow operator feature. It divides 36 regions of interest on the face and has good performance for subtle facial changes.

    \subsubsection{Deep learning MER methods}
    In addition to hand-crafted methods, deep neural networks have been widely used in MER. Due to the scarcity of ME data, unlike general neural networks which have a deep and complex structure to better extract features, networks in MER prefer to reduce the number of parameters to prevent overfitting. We selected OffApexNet\cite{Liong2018OFFApexNetOM}, STSTNet\cite{Liong2019ShallowTS}, RCN-A\cite{Xia2020RevealingTI}, MERSiam\cite{zhao2021two} and FR\cite{zhou2022feature} as four baseline methods. Off-ApexNet calculates optical flow maps between two frames and uses them to represent the entire video. STSTNet designs a shallow triple stream three-dimensional CNN that is computationally light whilst capable of extracting discriminative high-level features and details of MEs. RCN-A is a recurrent convolutional network (RCN) to explore the shallower-architecture and lower-resolution input data, shrinking model and input complexities simultaneously. MERSiam~\cite{zhao2021two} selects the optical flow sequences of ME videos as the model input and the method introduces a unique two-stage learning strategy, which includes prior-learning and target-learning stages. The main structure of MERSiam is built upon a Siamese 3D-CNN. FR aims to obtain salient and discriminative features for specific expressions and also predict expression by fusing the expression-specific features. It consists of an expression proposal module with the attention mechanism and a classification branch. 
    
    \subsection{Evaluation Implementation Settings}
    All experiments were conducted on 2 NVIDIA GeForce RTX 3090 GPUs with $2\times24$ GB memory. Following the original settings, the length of ME clips was 16 frames for 3D-CNN methods and the length of optical flow sequences was 10 for MERSiam. The spatial sizes of each input image or optical feature map were 224$\times$224 for R3D and I3D models, 28$\times$28 for OffApexNet, STSTNet and FR, 60$\times$60 for RCN-A and 112$\times$112 for MERSiam. 
    
    During training, cross-entropy loss and stochastic gradient descent ~(SGD) with a momentum of 0.9 were used to optimize the model parameters, and the batch size was set to 32 for 3D-CNN methods, 8 for MERSiam and 256 for other deep learning MER methods. Besides, other hyperparameters such as learning rate followed the settings in corresponding papers.
    
    \begin{figure*}[htbp]
    	\centering
            \subfigure[R3D] {\includegraphics[width=.33\textwidth]{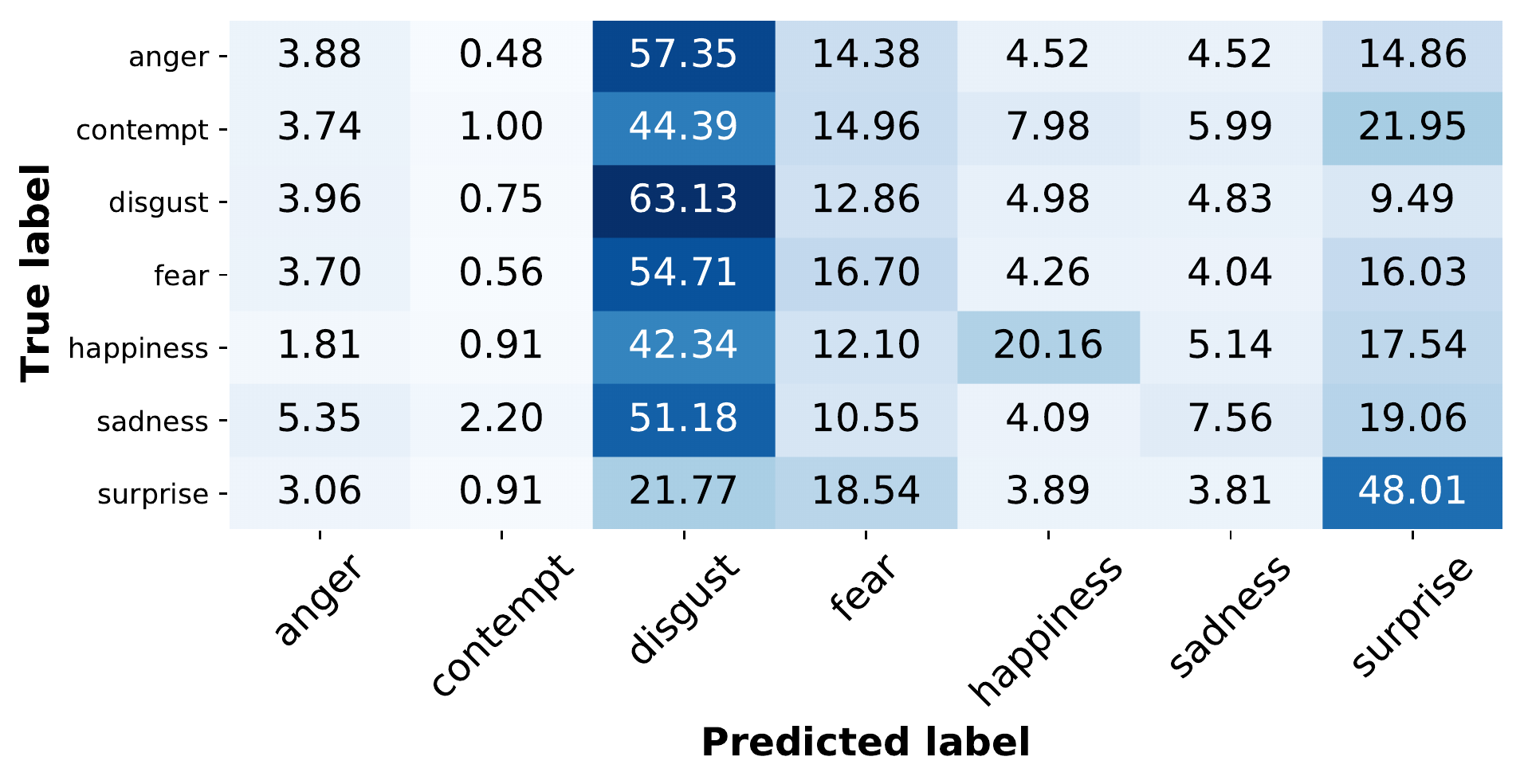}}
            \subfigure[I3D] {\includegraphics[width=.33\textwidth]{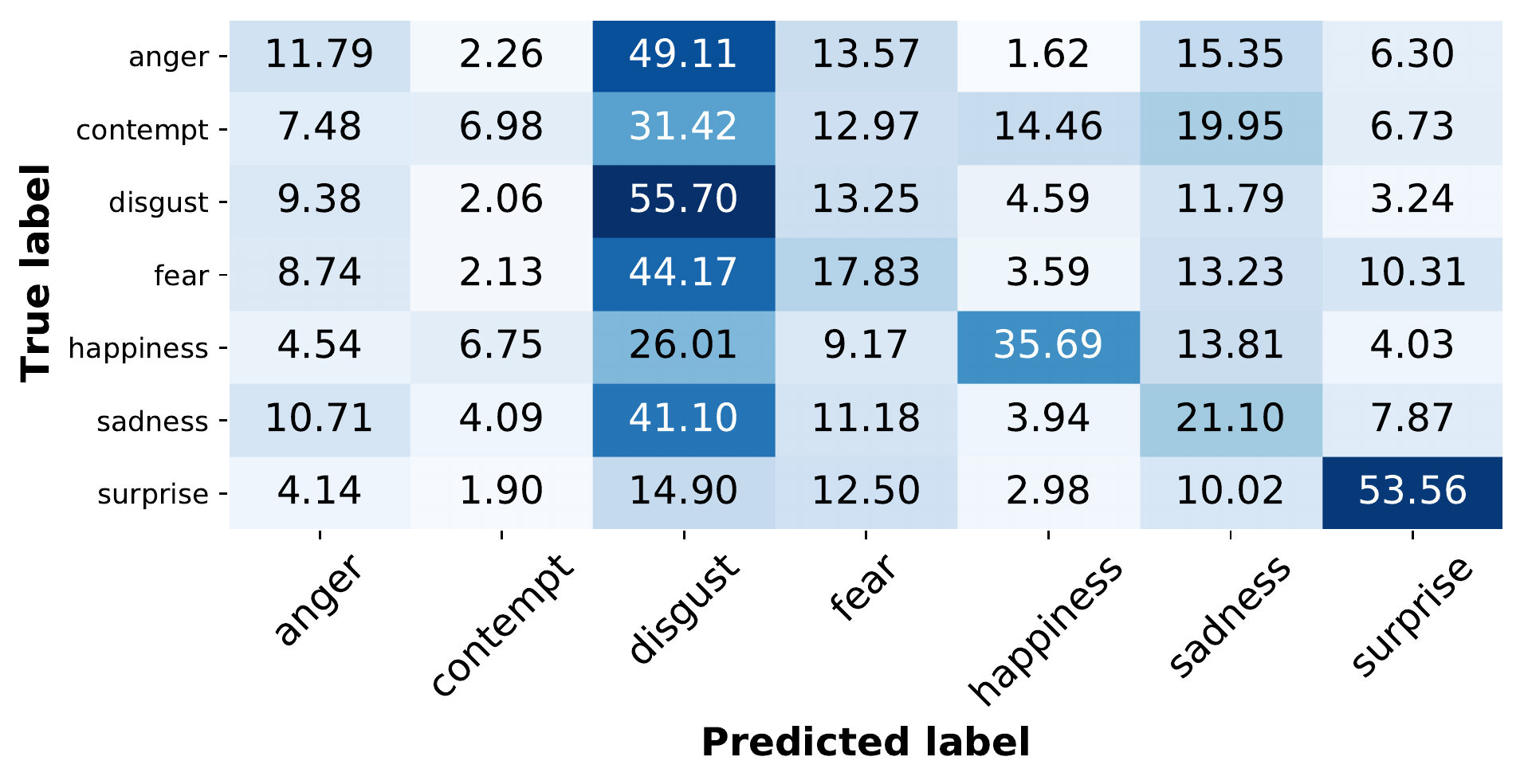}}
        	\subfigure[LBP-TOP] {\includegraphics[width=.33\textwidth]{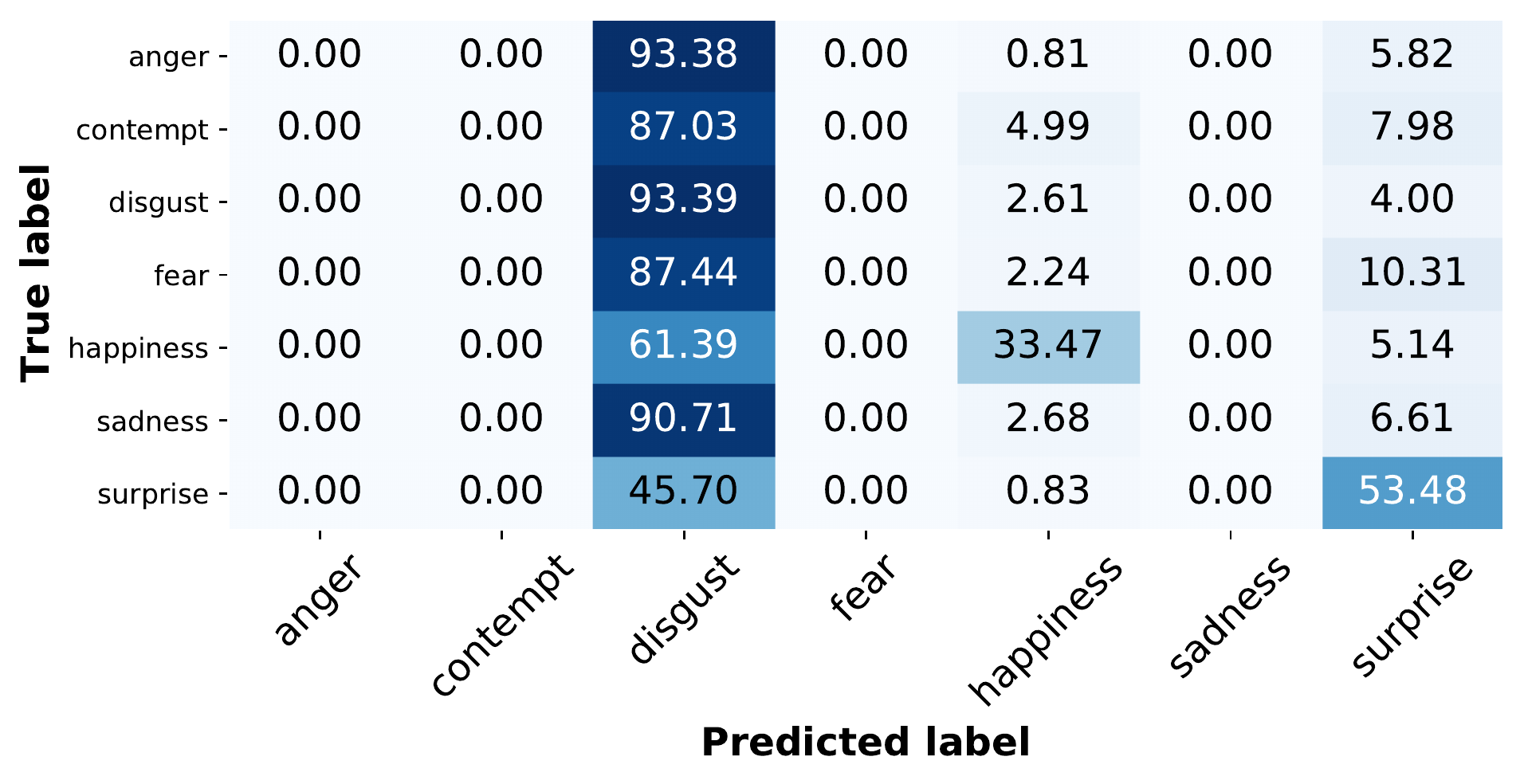}}
         
        	\subfigure[MDMO] {\includegraphics[width=.33\textwidth]{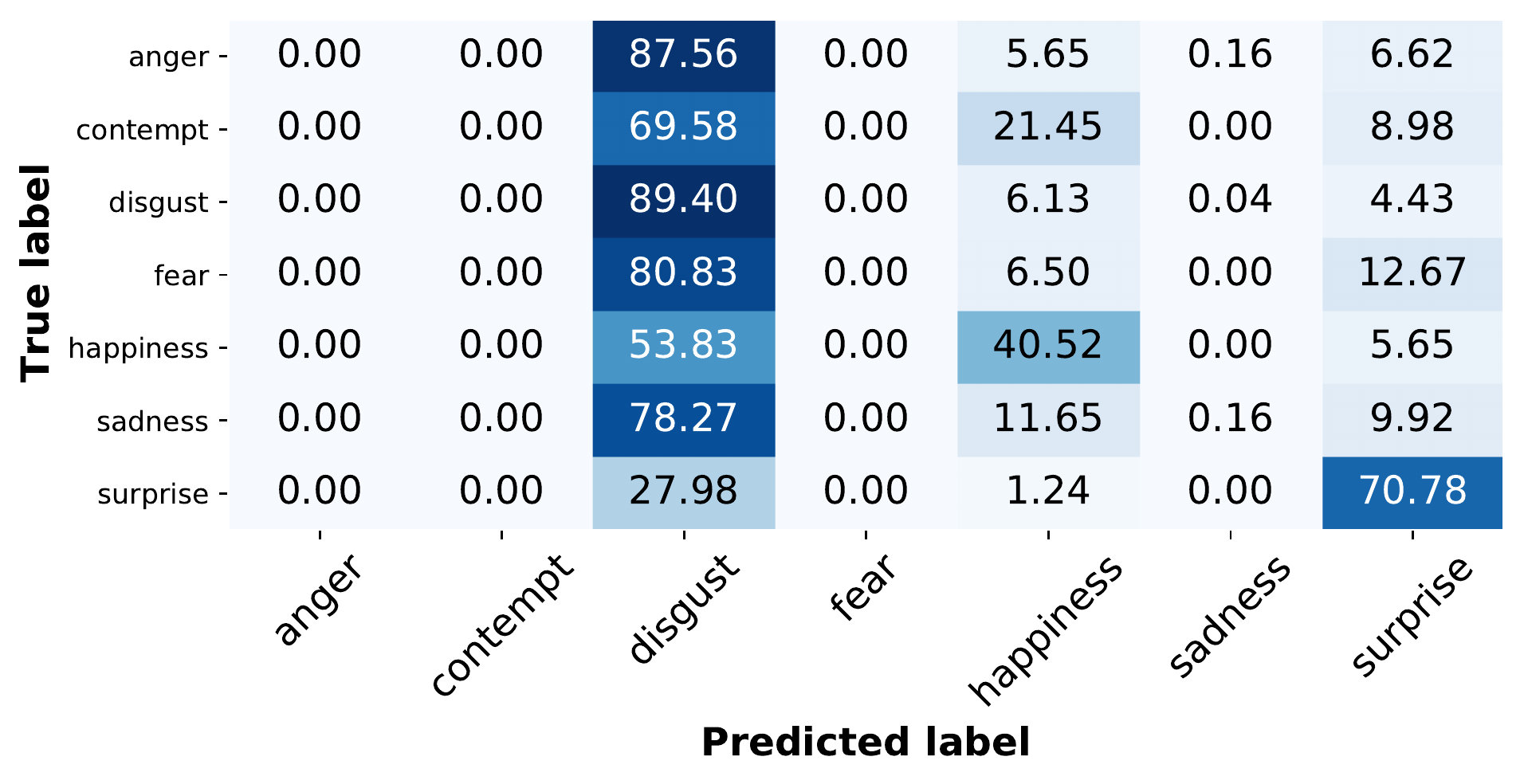}}
        	\subfigure[OffApexNet] {\includegraphics[width=.33\textwidth]{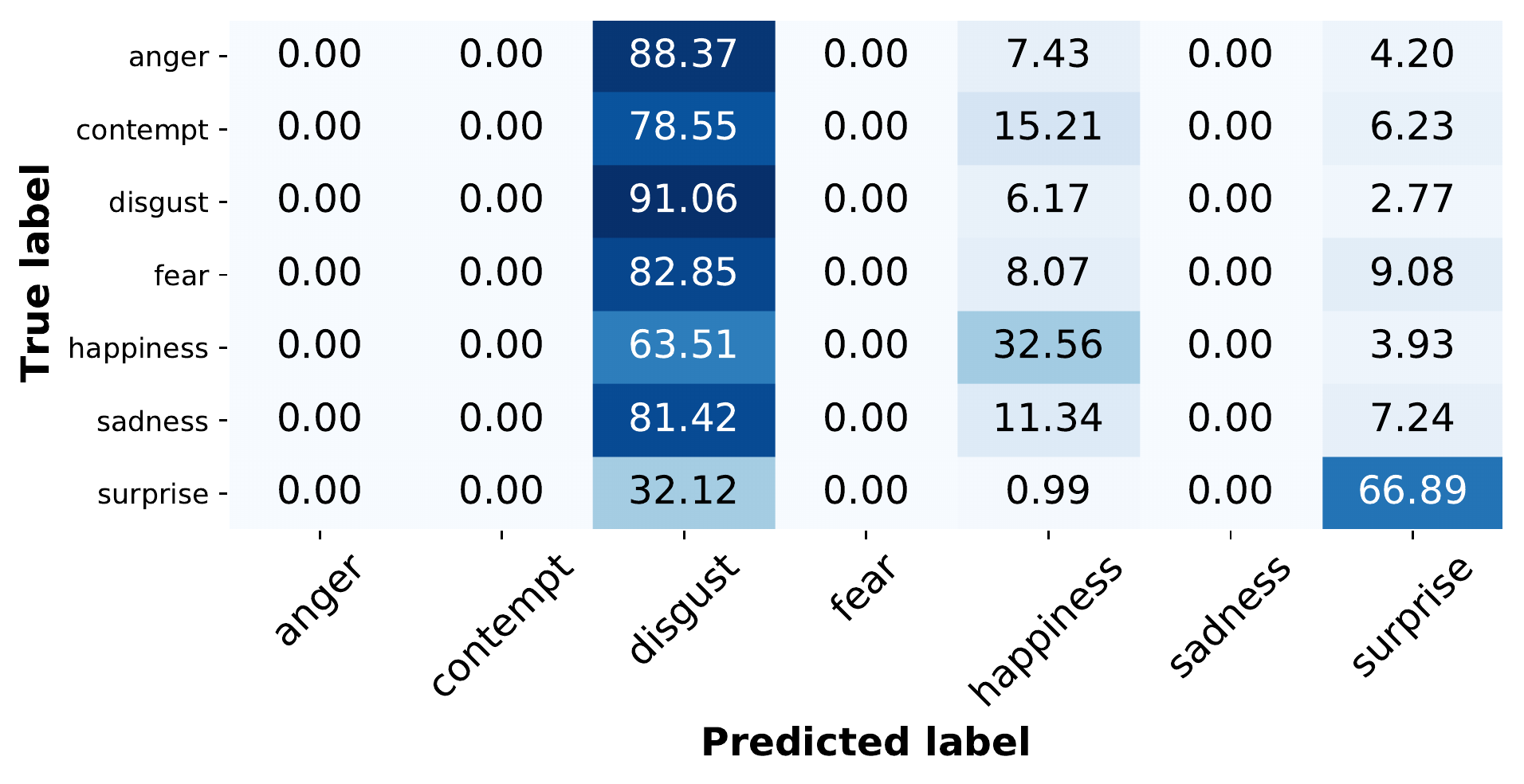}}
        	\subfigure[STSTNet] {\includegraphics[width=.33\textwidth]{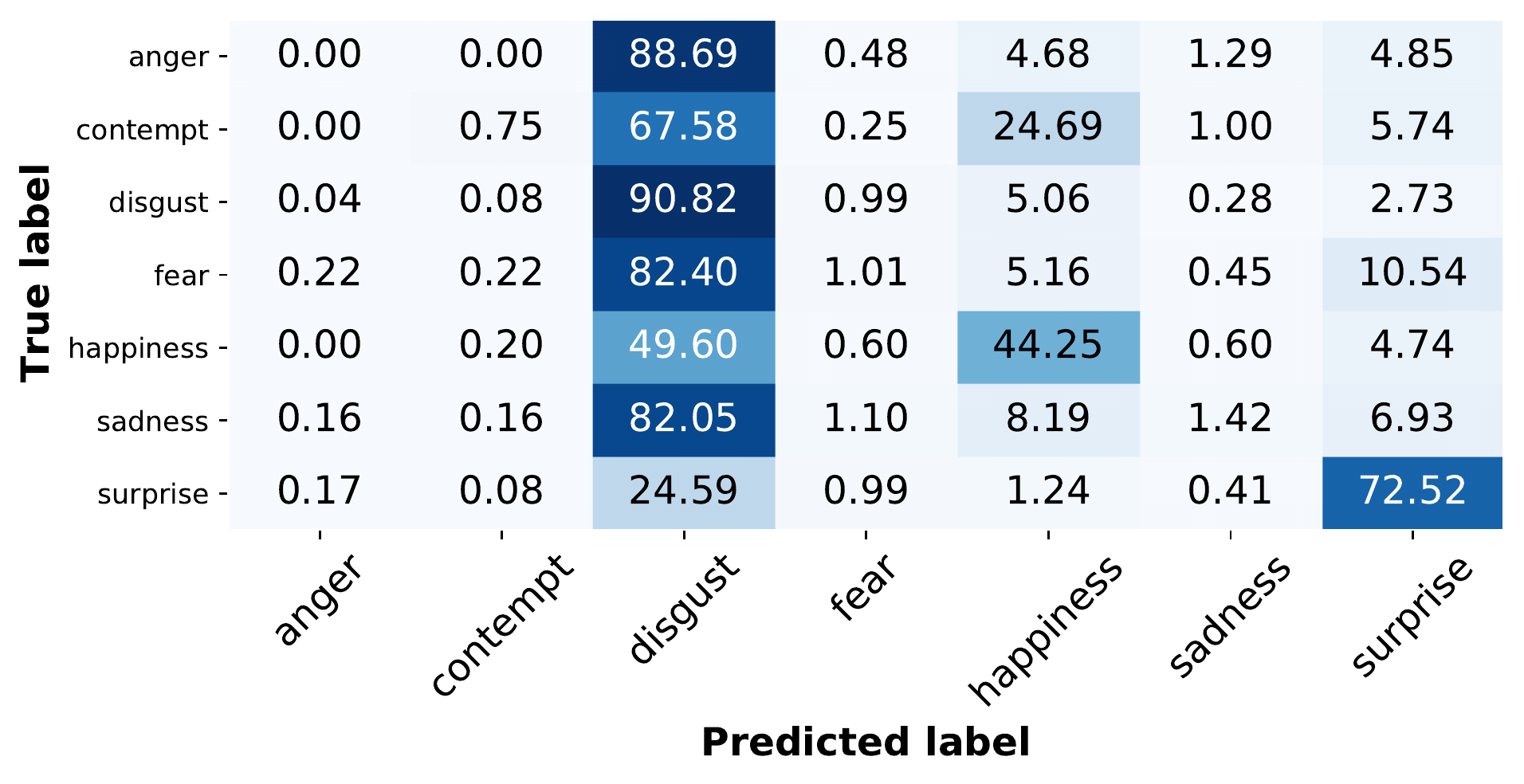}}
         
            \subfigure[RCN-A] {\includegraphics[width=.33\textwidth]{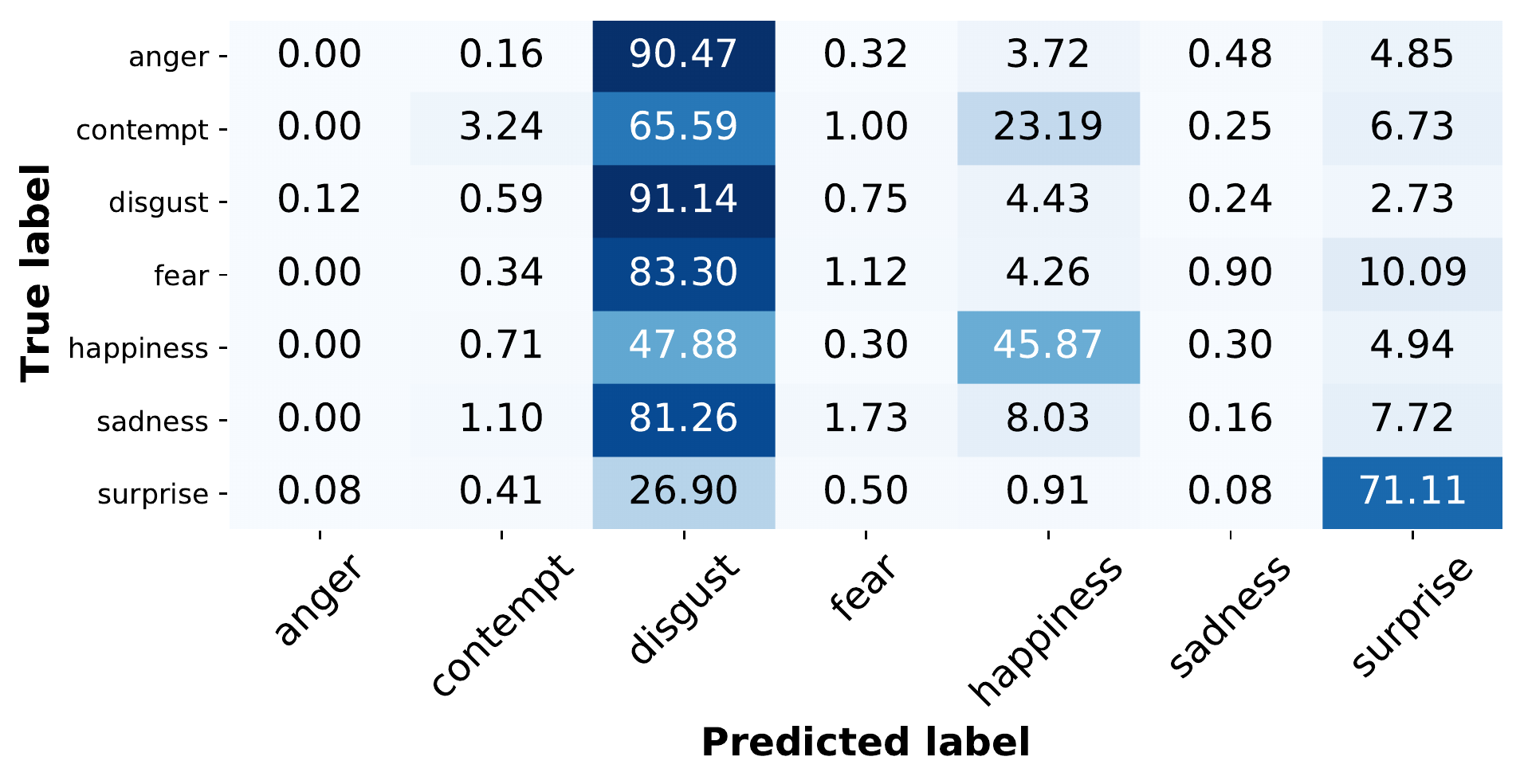}}
            \subfigure[MERSiam] {\includegraphics[width=.33\textwidth]{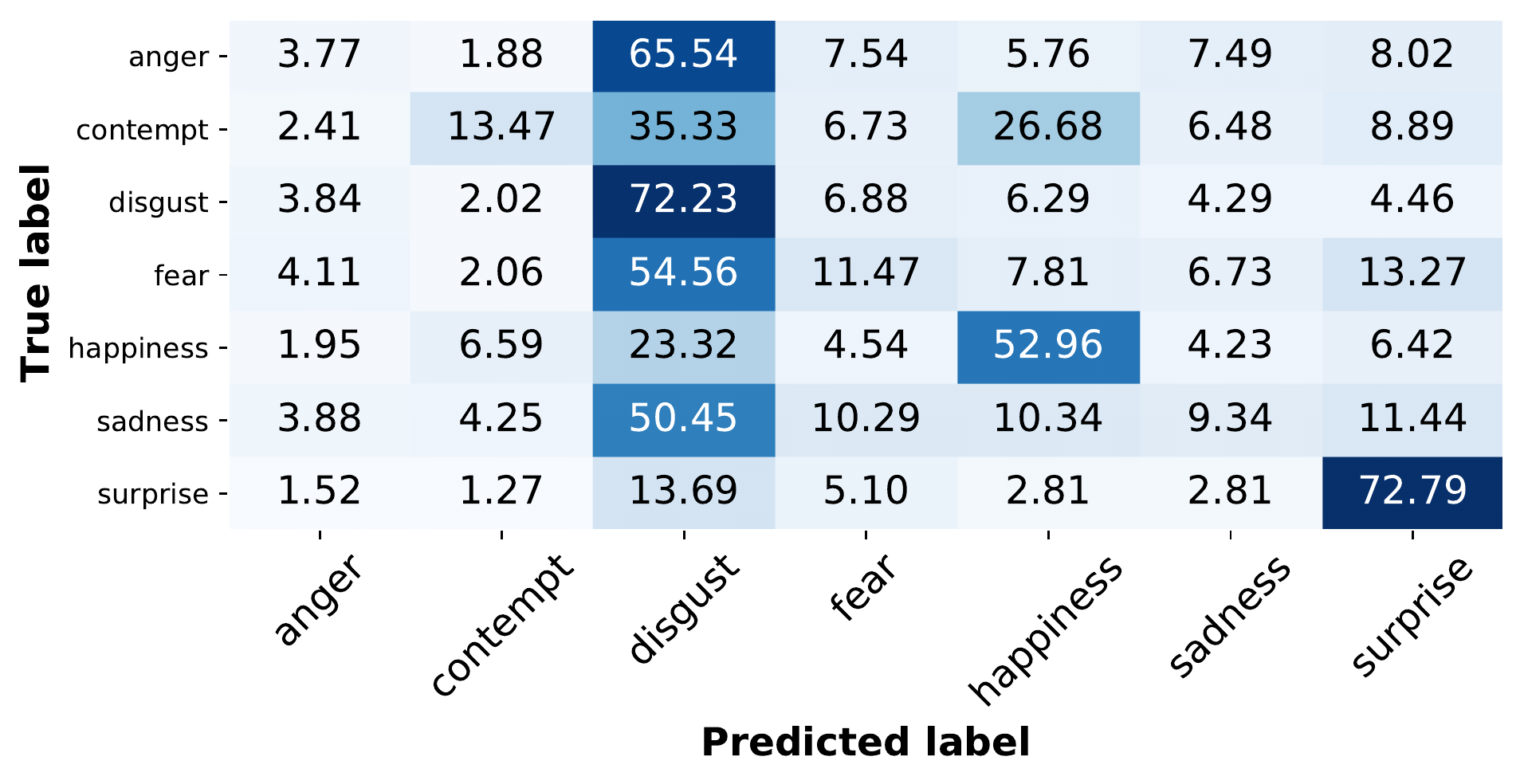}}
        	\subfigure[FR] {\includegraphics[width=.33\textwidth]{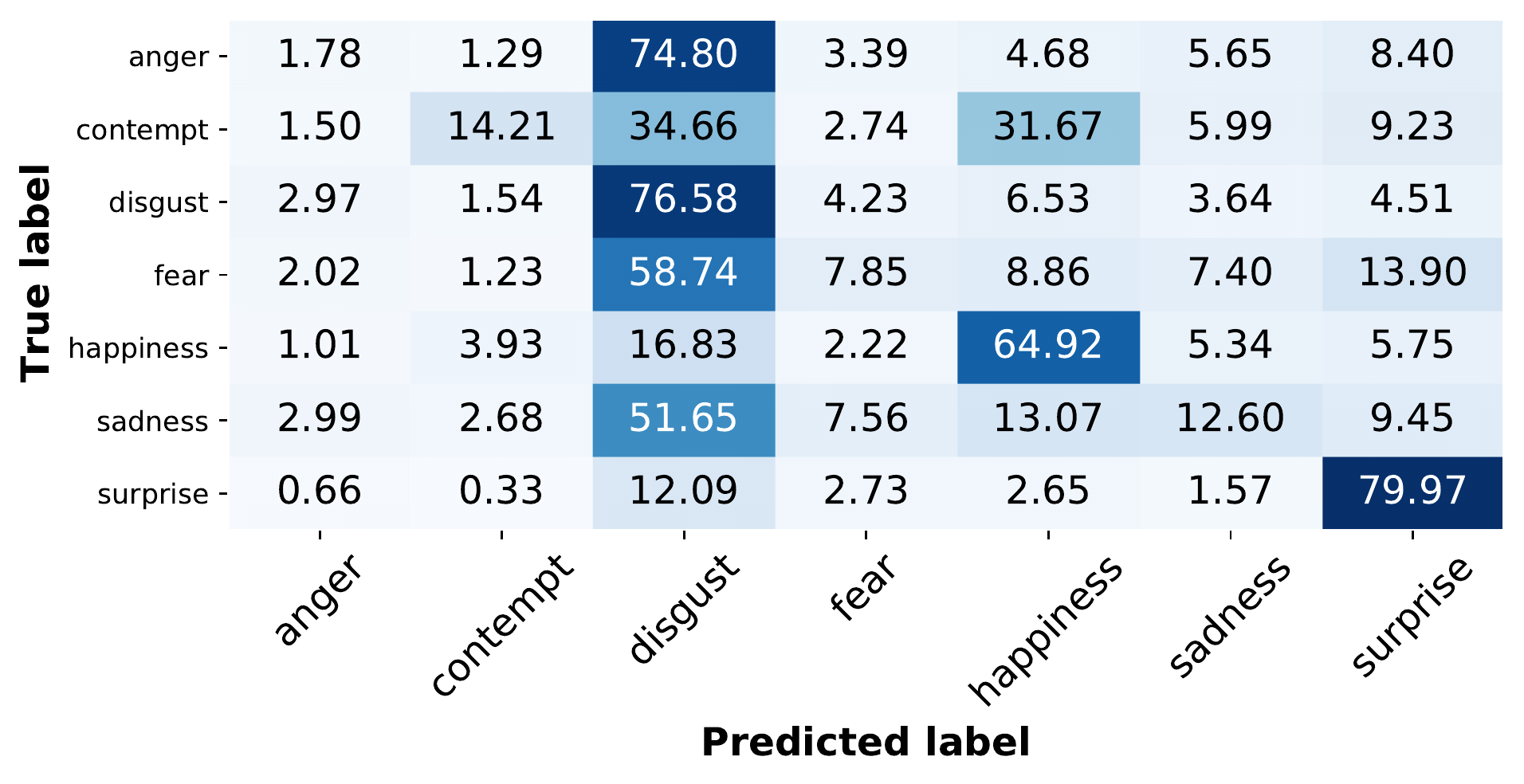}}
        	
    	\caption{Confusion matrices of baseline methods including 3D-CNN, hand-crafted MER and deep learning MER methods. }
    	\label{fig_E1}
    \end{figure*}
       
    \subsection{Evaluation Baseline Results}   
    \subsubsection{MER Results}
    To demonstrate the effectiveness of our DFME database for automatic MER tasks, we conducted a comprehensive MER experiment based on the above baseline methods. The results are shown in Table \ref{result}, and the recognition confusion matrix of each baseline model is shown in Fig.~\ref{fig_E1}. Comparing the MER results among the three groups of baseline methods, it's evident that most deep learning MER approaches exhibit better recognition performance compared to the others. This is consistent with our expectations because deep learning MER methods not only leverage the strengths of deep learning but also specifically design modules and introduce prior knowledge for ME features. Surprisingly, typical video understanding models like I3D have also achieved competitive recognition performance, surpassing two influential hand-crafted MER approaches. This reflects that by utilizing the DFME database we built, it's possible to train general video understanding models to acquire the ability for perceiving ME features.
    
    Furthermore, by observing the confusion matrices in Fig.~\ref{fig_E1}, we also discovered significant variations in the models' perception abilities for different categories of MEs. Taking FR as an example, it demonstrated higher recognition accuracy for \textit{disgust}, \textit{happiness}, and \textit{surprise} MEs, reaching 76.58\%, 64.92\%, and 79.97\%, respectively. However, the recognition accuracy for the remaining four ME categories were all below 15\%. The primary reasons contributing to this discrepancy can be identified as follows: 1) Negative emotions like \textit{anger, contempt, disgust, fear}, and \textit{sadness} exhibit significant similarities in ME. Referring to Table \ref{AU_percent}, it becomes apparent that MEs of \textit{anger, disgust} and \textit{fear} share common facial movements, including AU4, AU7, AU10, AU14, and AU24. 2) The problem of class imbalance persists within our DFME database. The number of \textit{disgust} samples (2528) is six times greater than \textit{contempt} (401)  which contains the least samples. To address these aspects, exploring more fine-grained ME feature learning models and solving the problem of biased ME feature learning caused by class imbalance remains key challenges in future MER studies.
    
    \begin{table}[htbp]
    \begin{threeparttable}    
    \centering
    \caption{ME recognition performance of baseline methods}\label{result}
    \renewcommand{\arraystretch}{1.2}
    \begin{tabular}{lllll}
    \hline
    Methods                                     &         & ACC(\%)        & UAR             & UF1             \\ \hline
    \multirow{2}{*}{\textit{3D-CNN Methods}}     & R3D\cite{Hara2018R3D}     & 36.62          & 0.2313          & 0.2164          \\
                                                & I3D\cite{Carreira2017I3D}     & 39.29          & 0.3058          & 0.2923          \\ \hline
    \multirow{2}{*}{\textit{Hand-crafted MER}}  & LBP-TOP\cite{zhao2007dynamic} & 46.82          & 0.2653          & 0.2336          \\
                                                & MDMO\cite{liu2015main}    & 49.34          & 0.2939          & 0.2489          \\ \hline
    \multirow{4}{*}{\textit{Deep Learning MER}} & OffApexNet\cite{Liong2018OFFApexNetOM} & 48.06          & 0.2806          & 0.2386          \\
                                                & STSTNet\cite{Liong2019ShallowTS} & 50.90          & 0.3108          & 0.2714          \\
                                                & RCN-A\cite{Xia2020RevealingTI}   & 50.98          & 0.3123          & 0.2751          \\
                                                & MERSiam\cite{zhao2021two}        &   52.18        &0.3532           & 0.3184          \\
                                                
                                                & FR\cite{zhou2022feature}       & \textbf{52.59} & \textbf{0.3814} & \textbf{0.3559} \\ \hline
    
    \end{tabular}
    \begin{tablenotes}[para,flushleft]  
     No data augmentation was used in any of the experiments.
    \end{tablenotes} 
    \end{threeparttable} 
    \end{table}

\subsubsection{AU Classification Results}
AU classification is crucial for analyzing facial behaviors. In this experiment, we conducted AU classification experiments on the DFME database using the aforementioned baseline methods. The experimental results are shown in Table \ref{resultAU}. It is evident that deep learning methods still achieve competitive results, especially FR method which performs best on this task. In addition, due to the large differences in the distribution of each AU in DFME, there are also significant differences in the classification performance of models for them. Overall, these models exhibit the best classification performance on common AUs such as AU1, AU2, and AU4. However, they tend to have a lower classification accuracy for low-frequency AU9 and AU15.

\begin{table*}[htbp]
\centering
\caption{AU classification performance of baseline methods}\label{resultAU}
\renewcommand{\arraystretch}{1.2}
\begin{tabular}{llllllllllllll}

\hline
Methods    & AU1    & AU2    & AU4    & AU5    & AU6    & AU7    & AU9    & AU10   & AU12   & AU14   & AU15   & AU17   & Average \\ \hline
R3D\cite{Hara2018R3D}        & 0.6061 & 0.5903 & 0.6967 & 0.4931 & 0.1878 & 0.3305 & 0.0000 & 0.0000 & 0.3102 & 0.0159 & 0.0000 & 0.0000 & 0.2692  \\
I3D\cite{Carreira2017I3D}        & 0.7517 & 0.7372 & 0.8082 & 0.6103 & 0.3221 & 0.4156 & 0.0000 & 0.1561 & 0.4697 & 0.3880 & 0.0000 & 0.3935 & 0.4210  \\ \hline
LBP-TOP\cite{zhao2007dynamic}    & 0.6282 & 0.5845 & 0.8297 & 0.3931 & 0.0722 & 0.1171 & 0.0000 & 0.0000 & 0.4007 & 0.0166 & 0.0000 & 0.0000 & 0.2535  \\
MDMO\cite{liu2015main}       & 0.5864 & 0.5783 & 0.7510 & 0.3736 & 0.0770 & 0.2079 & 0.0000 & 0.0082 & 0.2800 & 0.0897 & 0.0000 & 0.1675 & 0.2600  \\ \hline
OffApexNet\cite{Liong2018OFFApexNetOM} & 0.2973 & 0.0201 & 0.4367 & 0.0000 & 0.0000 & 0.0326 & 0.0229 & 0.1051 & 0.2805 & 0.0254  & 0.0410  & 0.0098 & 0.1060 \\
STSTNet\cite{Liong2019ShallowTS}    & 0.7256 & 0.6389 & 0.8833 & 0.3934 & 0.0567 & 0.1664 & 0.0000 & 0.0000 & 0.2654 & 0.0316 & 0.0000 & 0.0309 & 0.2660  \\
RCN-A\cite{Xia2020RevealingTI}      & 0.8120 & 0.8046 & 0.9076 & 0.5990 & 0.3478 & 0.4946 & 0.0588 & 0.1281 & 0.5157 & 0.1855 & 0.0000 & 0.2725 & 0.4272  \\
MERSiam\cite{zhao2021two} & 0.7710 & 0.7554 & 0.7306 & 0.5615 & 0.1383 & 0.2059 & 0.0000 & 0.1491 & 0.1774 & 0.0283 & 0.0000 & 0.3621 & 0.3233  \\
FR\cite{zhou2022feature}         & \textbf{0.8792} & \textbf{0.8837} & \textbf{0.9367} & \textbf{0.6971} & \textbf{0.4352} & \textbf{0.6217} & \textbf{0.0678} & \textbf{0.3373} & \textbf{0.7163} & \textbf{0.4638} & \textbf{0.3979} & \textbf{0.6605} & \textbf{0.5914}  \\ \hline
\end{tabular}
\end{table*}

\section{Conclusion and Future Work}
In this work, we focused on solving the problem of lacking abundant spontaneous ME data for MER. To this end, we built a new ME database called DFME containing 7,526 ME videos spanning multiple high frame rates. To the best of our knowledge, DFME has the largest ME sample size at present. Furthermore, to verify the feasibility and validity of DFME database for MER and AU classification tasks, we reproduced some spatiotemporal visual feature learning models and hand-crafted and deep learning MER methods on DFME database, objectively verifying the reliability of data quality, and providing a benchmark for subsequent MER studies. 

In the future, we will strive to expand the DFME database to provide more abundant ME data for automatic ME analysis research, including the collection of multimodal ME data in multiple natural scenes. Based on this, we will also study the high accuracy and robust MER models, such as self-supervised MER combined with more samples with uncertain labels, and apply them to actual scenes.


%

\ifCLASSOPTIONcompsoc
  \section*{Acknowledgments}
\else
  \section*{Acknowledgment}
\fi
This work has received a lot of guidance and help from the teachers in the Micro-expression Laboratory of Institute of Psychology, Chinese Academy of Sciences. We would like to express our special thanks to them. 

\ifCLASSOPTIONcaptionsoff
  \newpage
\fi


\bibliographystyle{IEEEtran}
%
\bibliography{Reference}

\begin{thebibliography}{10}
\providecommand{\url}[1]{#1}
\csname url@samestyle\endcsname
\providecommand{\newblock}{\relax}
\providecommand{\bibinfo}[2]{#2}
\providecommand{\BIBentrySTDinterwordspacing}{\spaceskip=0pt\relax}
\providecommand{\BIBentryALTinterwordstretchfactor}{4}
\providecommand{\BIBentryALTinterwordspacing}{\spaceskip=\fontdimen2\font plus
\BIBentryALTinterwordstretchfactor\fontdimen3\font minus \fontdimen4\font\relax}
\providecommand{\BIBforeignlanguage}[2]{{%
\expandafter\ifx\csname l@#1\endcsname\relax
\typeout{** WARNING: IEEEtran.bst: No hyphenation pattern has been}%
\typeout{** loaded for the language `#1'. Using the pattern for}%
\typeout{** the default language instead.}%
\else
\language=\csname l@#1\endcsname
\fi
#2}}
\providecommand{\BIBdecl}{\relax}
\BIBdecl

\bibitem{mehrabian2017communication}
A.~Mehrabian, ``Communication without words,'' in \emph{Communication theory}.\hskip 1em plus 0.5em minus 0.4em\relax Routledge, 2017, pp. 193--200.

\bibitem{Zhao2023FacialMA}
\BIBentryALTinterwordspacing
G.~Zhao, X.~Li, Y.~Li, and M.~Pietik{\"a}inen, ``Facial micro-expressions: An overview,'' \emph{Proceedings of the IEEE}, 2023. [Online]. Available: \url{https://api.semanticscholar.org/CorpusID:259823625}
\BIBentrySTDinterwordspacing

\bibitem{haggard1966micromomentary}
E.~A. Haggard and K.~S. Isaacs, ``Micromomentary facial expressions as indicators of ego mechanisms in psychotherapy,'' in \emph{Methods of research in psychotherapy}.\hskip 1em plus 0.5em minus 0.4em\relax Springer, 1966, pp. 154--165.

\bibitem{ekman1969nonverbal}
P.~Ekman and W.~V. Friesen, ``Nonverbal leakage and clues to deception,'' \emph{Psychiatry}, vol.~32, no.~1, pp. 88--106, 1969.

\bibitem{porter2008reading}
S.~Porter and L.~Ten~Brinke, ``Reading between the lies: Identifying concealed and falsified emotions in universal facial expressions,'' \emph{Psychological science}, vol.~19, no.~5, pp. 508--514, 2008.

\bibitem{ekman2009telling}
P.~Ekman, \emph{Telling lies: Clues to deceit in the marketplace, politics, and marriage (revised edition)}.\hskip 1em plus 0.5em minus 0.4em\relax WW Norton \& Company, 2009.

\bibitem{weinberger2010intent}
S.~Weinberger, ``Intent to deceive? can the science of deception detection help to catch terrorists? sharon weinberger takes a close look at the evidence for it,'' \emph{Nature}, vol. 465, no. 7297, pp. 412--416, 2010.

\bibitem{hunter2020emotional}
L.~Hunter, L.~Roland, and A.~Ferozpuri, ``Emotional expression processing and depressive symptomatology: Eye-tracking reveals differential importance of lower and middle facial areas of interest,'' \emph{Depression Research and Treatment}, vol. 2020, 2020.

\bibitem{microreactions}
J.~Zhenyu, \emph{Micro-reactions}.\hskip 1em plus 0.5em minus 0.4em\relax Beijing: China Friendship Publishing Company, 2020.

\bibitem{Lombardi2017PsychologicalSD}
\BIBentryALTinterwordspacing
L.~Lombardi and F.~Marcolin, ``Psychological stress detection by 2d and 3d facial image processing,'' in \emph{Intelligent Information Hiding and Multimedia Signal Processing}, 2017. [Online]. Available: \url{https://api.semanticscholar.org/CorpusID:222112061}
\BIBentrySTDinterwordspacing

\bibitem{wang2021single}
J.~Wang, X.~Pan, X.~Li, G.~Wei, and Y.~Zhou, ``Single trunk multi-scale network for micro-expression recognition,'' \emph{Graphics and Visual Computing}, vol.~4, p. 200026, 2021.

\bibitem{ekman2003micro}
P.~Ekman, ``Micro expressions training tool,'' \url={Emotionsrevealed.com}, 2003.

\bibitem{Russell2006APS}
\BIBentryALTinterwordspacing
T.~A. Russell, E.~M.-Y. Chu, and M.~L. Phillips, ``A pilot study to investigate the effectiveness of emotion recognition remediation in schizophrenia using the micro-expression training tool.'' \emph{The British journal of clinical psychology}, vol. 45 Pt 4, pp. 579--83, 2006. [Online]. Available: \url{https://api.semanticscholar.org/CorpusID:23119588}
\BIBentrySTDinterwordspacing

\bibitem{Endres2009MicroexpressionRT}
\BIBentryALTinterwordspacing
J.~Endres and A.~H. Laidlaw, ``Micro-expression recognition training in medical students: a pilot study,'' \emph{BMC Medical Education}, vol.~9, pp. 47 -- 47, 2009. [Online]. Available: \url{https://api.semanticscholar.org/CorpusID:18933437}
\BIBentrySTDinterwordspacing

\bibitem{frank2009see}
M.~Frank, M.~Herbasz, K.~Sinuk, A.~Keller, and C.~Nolan, ``I see how you feel: Training laypeople and professionals to recognize fleeting emotions,'' in \emph{The annual meeting of the international communication association. Sheraton New York, New York City}, 2009, pp. 1--35.

\bibitem{pfister2011recognising}
T.~Pfister, X.~Li, G.~Zhao, and M.~Pietik{\"a}inen, ``Recognising spontaneous facial micro-expressions,'' in \emph{2011 international conference on computer vision}.\hskip 1em plus 0.5em minus 0.4em\relax IEEE, 2011, pp. 1449--1456.

\bibitem{zhao2007dynamic}
G.~Zhao and M.~Pietikainen, ``Dynamic texture recognition using local binary patterns with an application to facial expressions,'' \emph{IEEE Transactions on Pattern Analysis \& Machine Intelligence}, no.~6, pp. 915--928, 2007.

\bibitem{wang2015lbp}
Y.~Wang, J.~See, R.~C.-W. Phan, and Y.-H. Oh, ``Lbp with six intersection points: Reducing redundant information in lbp-top for micro-expression recognition,'' in \emph{Computer Vision--ACCV 2014: 12th Asian Conference on Computer Vision, Singapore, Singapore, November 1-5, 2014, Revised Selected Papers, Part I 12}.\hskip 1em plus 0.5em minus 0.4em\relax Springer, 2015, pp. 525--537.

\bibitem{huang2016spontaneous}
X.~Huang, G.~Zhao, X.~Hong, W.~Zheng, and M.~Pietik{\"a}inen, ``Spontaneous facial micro-expression analysis using spatiotemporal completed local quantized patterns,'' \emph{Neurocomputing}, vol. 175, pp. 564--578, 2016.

\bibitem{liu2015main}
Y.-J. Liu, J.-K. Zhang, W.-J. Yan, S.-J. Wang, G.~Zhao, and X.~Fu, ``A main directional mean optical flow feature for spontaneous micro-expression recognition,'' \emph{IEEE Transactions on Affective Computing}, vol.~7, no.~4, pp. 299--310, 2015.

\bibitem{wang2018micro}
S.-J. Wang, B.-J. Li, Y.-J. Liu, W.-J. Yan, X.~Ou, X.~Huang, F.~Xu, and X.~Fu, ``Micro-expression recognition with small sample size by transferring long-term convolutional neural network,'' \emph{Neurocomputing}, vol. 312, pp. 251--262, 2018.

\bibitem{zhao2021two}
S.~Zhao, H.~Tao, Y.~Zhang, T.~Xu, K.~Zhang, Z.~Hao, and E.~Chen, ``A two-stage 3d cnn based learning method for spontaneous micro-expression recognition,'' \emph{Neurocomputing}, vol. 448, pp. 276--289, 2021.

\bibitem{9854172}
Q.~Mao, L.~Zhou, W.~Zheng, X.~Shao, and X.~Huang, ``Objective class-based micro-expression recognition under partial occlusion via region-inspired relation reasoning network,'' \emph{IEEE Transactions on Affective Computing}, vol.~13, no.~4, pp. 1998--2016, 2022.

\bibitem{ben2021video}
X.~Ben, Y.~Ren, J.~Zhang, S.-J. Wang, K.~Kpalma, W.~Meng, and Y.-J. Liu, ``Video-based facial micro-expression analysis: A survey of datasets, features and algorithms,'' \emph{IEEE transactions on pattern analysis and machine intelligence}, 2021.

\bibitem{Li2021DeepLF}
\BIBentryALTinterwordspacing
Y.~Li, J.~Wei, Y.~Liu, J.~Kauttonen, and G.~Zhao, ``Deep learning for micro-expression recognition: A survey,'' \emph{IEEE Transactions on Affective Computing}, vol.~13, pp. 2028--2046, 2021. [Online]. Available: \url{https://api.semanticscholar.org/CorpusID:237532789}
\BIBentrySTDinterwordspacing

\bibitem{Xia2020LearningFM}
\BIBentryALTinterwordspacing
B.~Xia, W.~Wang, S.~Wang, and E.~Chen, ``Learning from macro-expression: a micro-expression recognition framework,'' \emph{Proceedings of the 28th ACM International Conference on Multimedia}, 2020. [Online]. Available: \url{https://api.semanticscholar.org/CorpusID:222278587}
\BIBentrySTDinterwordspacing

\bibitem{li2013spontaneous}
X.~Li, T.~Pfister, X.~Huang, G.~Zhao, and M.~Pietik{\"a}inen, ``A spontaneous micro-expression database: Inducement, collection and baseline,'' in \emph{2013 10th IEEE International Conference and Workshops on Automatic face and gesture recognition (fg)}.\hskip 1em plus 0.5em minus 0.4em\relax IEEE, 2013, pp. 1--6.

\bibitem{yan2014casme}
W.-J. Yan, X.~Li, S.-J. Wang, G.~Zhao, Y.-J. Liu, Y.-H. Chen, and X.~Fu, ``Casme ii: An improved spontaneous micro-expression database and the baseline evaluation,'' \emph{PloS one}, vol.~9, no.~1, p. e86041, 2014.

\bibitem{davison2016samm}
A.~K. Davison, C.~Lansley, N.~Costen, K.~Tan, and M.~H. Yap, ``Samm: A spontaneous micro-facial movement dataset,'' \emph{IEEE transactions on affective computing}, vol.~9, no.~1, pp. 116--129, 2016.

\bibitem{li2022cas}
J.~Li, Z.~Dong, S.~Lu, S.-J. Wang, W.-J. Yan, Y.~Ma, Y.~Liu, C.~Huang, and X.~Fu, ``Cas (me) 3: A third generation facial spontaneous micro-expression database with depth information and high ecological validity,'' \emph{IEEE Transactions on Pattern Analysis and Machine Intelligence}, 2022.

\bibitem{MEGC2022}
\BIBentryALTinterwordspacing
J.~Li, M.~H. Yap, W.-H. Cheng, J.~See, X.~Hong, X.~Li, S.-J. Wang, A.~K. Davison, Y.~Li, and Z.~Dong, ``Megc2022: Acm multimedia 2022 micro-expression grand challenge,'' in \emph{Proceedings of the 30th ACM International Conference on Multimedia}, ser. MM '22.\hskip 1em plus 0.5em minus 0.4em\relax New York, NY, USA: Association for Computing Machinery, 2022, p. 7170–7174. [Online]. Available: \url{https://doi.org/10.1145/3503161.3551601}
\BIBentrySTDinterwordspacing

\bibitem{xu2021famgan}
Y.~Xu, S.~Zhao, H.~Tang, X.~Mao, T.~Xu, and E.~Chen, ``Famgan: Fine-grained aus modulation based generative adversarial network for micro-expression generation,'' in \emph{Proceedings of the 29th ACM International Conference on Multimedia}, 2021, pp. 4813--4817.

\bibitem{zhao2022fine}
S.~Zhao, S.~Yin, H.~Tang, R.~Jin, Y.~Xu, T.~Xu, and E.~Chen, ``Fine-grained micro-expression generation based on thin-plate spline and relative au constraint,'' in \emph{Proceedings of the 30th ACM International Conference on Multimedia}, 2022, pp. 7150--7154.

\bibitem{shreve2011macro}
M.~Shreve, S.~Godavarthy, D.~Goldgof, and S.~Sarkar, ``Macro-and micro-expression spotting in long videos using spatio-temporal strain,'' in \emph{2011 IEEE International Conference on Automatic Face \& Gesture Recognition (FG)}.\hskip 1em plus 0.5em minus 0.4em\relax IEEE, 2011, pp. 51--56.

\bibitem{polikovsky2009facial}
S.~Polikovsky, Y.~Kameda, and Y.~Ohta, ``Facial micro-expressions recognition using high speed camera and 3d-gradient descriptor,'' in \emph{3rd International Conference on Imaging for Crime Detection and Prevention (ICDP)}, 2009, pp. 1--6.

\bibitem{yan2013casme}
W.-J. Yan, Q.~Wu, Y.-J. Liu, S.-J. Wang, and X.~Fu, ``Casme database: A dataset of spontaneous micro-expressions collected from neutralized faces,'' in \emph{2013 10th IEEE international conference and workshops on automatic face and gesture recognition (FG)}.\hskip 1em plus 0.5em minus 0.4em\relax IEEE, 2013, pp. 1--7.

\bibitem{qu2017cas}
F.~Qu, S.-J. Wang, W.-J. Yan, H.~Li, S.~Wu, and X.~Fu, ``Cas(me)$^{2}$: a database for spontaneous macro-expression and micro-expression spotting and recognition,'' \emph{IEEE Transactions on Affective Computing}, vol.~9, no.~4, pp. 424--436, 2017.

\bibitem{li20224dme}
X.~Li, S.~Cheng, Y.~Li, M.~Behzad, J.~Shen, S.~Zafeiriou, M.~Pantic, and G.~Zhao, ``4dme: A spontaneous 4d micro-expression dataset with multimodalities,'' \emph{IEEE Transactions on Affective Computing}, 2022.

\bibitem{husak2017spotting}
P.~Hus{\'a}k, J.~Cech, and J.~Matas, ``Spotting facial micro-expressions “in the wild”,'' in \emph{22nd Computer Vision Winter Workshop (Retz)}, 2017, pp. 1--9.

\bibitem{Kumar2021MicroExpressionCB}
A.~J.~R. Kumar and B.~Bhanu, ``Micro-expression classification based on landmark relations with graph attention convolutional network,'' \emph{2021 IEEE/CVF Conference on Computer Vision and Pattern Recognition Workshops (CVPRW)}, pp. 1511--1520, 2021.

\bibitem{takalkar2021lgattnet}
M.~A. Takalkar, S.~Thuseethan, S.~Rajasegarar, Z.~Chaczko, M.~Xu, and J.~Yearwood, ``Lgattnet: Automatic micro-expression detection using dual-stream local and global attentions,'' \emph{Knowledge-Based Systems}, vol. 212, p. 106566, 2021.

\bibitem{pan2021micro}
H.~Pan, L.~Xie, J.~Li, Z.~Lv, and Z.~Wang, ``Micro-expression recognition by two-stream difference network,'' \emph{IET Computer Vision}, vol.~15, no.~6, pp. 440--448, 2021.

\bibitem{yap2018facial}
M.~H. Yap, J.~See, X.~Hong, and S.-J. Wang, ``Facial micro-expressions grand challenge 2018 summary,'' in \emph{2018 13th IEEE International Conference on Automatic Face \& Gesture Recognition (FG 2018)}.\hskip 1em plus 0.5em minus 0.4em\relax IEEE, 2018, pp. 675--678.

\bibitem{see2019megc}
J.~See, M.~H. Yap, J.~Li, X.~Hong, and S.-J. Wang, ``Megc 2019--the second facial micro-expressions grand challenge,'' in \emph{2019 14th IEEE International Conference on Automatic Face \& Gesture Recognition (FG 2019)}.\hskip 1em plus 0.5em minus 0.4em\relax IEEE, 2019, pp. 1--5.

\bibitem{zong2019cross}
Y.~Zong, W.~Zheng, X.~Hong, C.~Tang, Z.~Cui, and G.~Zhao, ``Cross-database micro-expression recognition: A benchmark,'' in \emph{Proceedings of the 2019 on International Conference on Multimedia Retrieval}, 2019, pp. 354--363.

\bibitem{chaudhry2009histograms}
R.~Chaudhry, A.~Ravichandran, G.~Hager, and R.~Vidal, ``Histograms of oriented optical flow and binet-cauchy kernels on nonlinear dynamical systems for the recognition of human actions,'' in \emph{2009 IEEE Conference on Computer Vision and Pattern Recognition}.\hskip 1em plus 0.5em minus 0.4em\relax IEEE, 2009, pp. 1932--1939.

\bibitem{huang2017discriminative}
X.~Huang, S.-J. Wang, X.~Liu, G.~Zhao, X.~Feng, and M.~Pietik{\"a}inen, ``Discriminative spatiotemporal local binary pattern with revisited integral projection for spontaneous facial micro-expression recognition,'' \emph{IEEE Transactions on Affective Computing}, vol.~10, no.~1, pp. 32--47, 2019.

\bibitem{li2017towards}
X.~Li, X.~Hong, A.~Moilanen, X.~Huang, T.~Pfister, G.~Zhao, and M.~Pietik{\"a}inen, ``Towards reading hidden emotions: A comparative study of spontaneous micro-expression spotting and recognition methods,'' \emph{IEEE transactions on affective computing}, vol.~9, no.~4, pp. 563--577, 2018.

\bibitem{xu2017microexpression}
F.~Xu, J.~Zhang, and J.~Z. Wang, ``Microexpression identification and categorization using a facial dynamics map,'' \emph{IEEE Transactions on Affective Computing}, vol.~8, no.~2, pp. 254--267, 2017.

\bibitem{peng2018macro}
M.~Peng, Z.~Wu, Z.~Zhang, and T.~Chen, ``From macro to micro expression recognition: Deep learning on small datasets using transfer learning,'' in \emph{2018 13th IEEE International Conference on Automatic Face \& Gesture Recognition (FG 2018)}.\hskip 1em plus 0.5em minus 0.4em\relax IEEE, 2018, pp. 657--661.

\bibitem{he2016deep}
K.~He, X.~Zhang, S.~Ren, and J.~Sun, ``Deep residual learning for image recognition,'' in \emph{Proceedings of the IEEE conference on computer vision and pattern recognition}, 2016, pp. 770--778.

\bibitem{van2019capsulenet}
N.~Van~Quang, J.~Chun, and T.~Tokuyama, ``Capsulenet for micro-expression recognition,'' in \emph{2019 14th IEEE International Conference on Automatic Face \& Gesture Recognition (FG 2019)}.\hskip 1em plus 0.5em minus 0.4em\relax IEEE, 2019, pp. 1--7.

\bibitem{xia2020learning}
B.~Xia, W.~Wang, S.~Wang, and E.~Chen, ``Learning from macro-expression: a micro-expression recognition framework,'' in \emph{Proceedings of the 28th ACM International Conference on Multimedia}, 2020, pp. 2936--2944.

\bibitem{li2020joint}
Y.~Li, X.~Huang, and G.~Zhao, ``Joint local and global information learning with single apex frame detection for micro-expression recognition,'' \emph{IEEE Transactions on Image Processing}, vol.~30, pp. 249--263, 2020.

\bibitem{liong2018less}
S.-T. Liong, J.~See, K.~Wong, and R.~C.-W. Phan, ``Less is more: Micro-expression recognition from video using apex frame,'' \emph{Signal Processing: Image Communication}, vol.~62, pp. 82--92, 2018.

\bibitem{liu2019neural}
Y.~Liu, H.~Du, L.~Zheng, and T.~Gedeon, ``A neural micro-expression recognizer,'' in \emph{2019 14th IEEE International Conference on Automatic Face \& Gesture Recognition (FG 2019)}.\hskip 1em plus 0.5em minus 0.4em\relax IEEE, 2019, pp. 1--4.

\bibitem{zhou2022feature}
L.~Zhou, Q.~Mao, X.~Huang, F.~Zhang, and Z.~Zhang, ``Feature refinement: An expression-specific feature learning and fusion method for micro-expression recognition,'' \emph{Pattern Recognition}, vol. 122, p. 108275, 2022.

\bibitem{Gong2022MetaMMFNetMB}
W.~Gong, Y.~Zhang, W.~Wang, P.~Cheng, and J.~Gonz{\`a}lez, ``Meta-mmfnet: Meta-learning based multi-model fusion network for micro-expression recognition,'' \emph{ACM Transactions on Multimedia Computing, Communications, and Applications (TOMM)}, 2022.

\bibitem{Liu2022MicroexpressionRB}
S.~Liu, Y.~Ren, L.~Li, X.~Sun, Y.~Song, and C.-C. Hung, ``Micro-expression recognition based on squeezenet and c3d,'' \emph{Multimedia Systems}, vol.~28, pp. 2227 -- 2236, 2022.

\bibitem{Iandola2016SqueezeNetAA}
F.~N. Iandola, M.~W. Moskewicz, K.~Ashraf, S.~Han, W.~J. Dally, and K.~Keutzer, ``Squeezenet: Alexnet-level accuracy with 50x fewer parameters and <1mb model size,'' \emph{ArXiv}, vol. abs/1602.07360, 2016.

\bibitem{kim2016micro}
D.~H. Kim, W.~J. Baddar, and Y.~M. Ro, ``Micro-expression recognition with expression-state constrained spatio-temporal feature representations,'' in \emph{Proceedings of the 24th ACM international conference on Multimedia}.\hskip 1em plus 0.5em minus 0.4em\relax ACM, 2016, pp. 382--386.

\bibitem{khor2018enriched}
H.-Q. Khor, J.~See, R.~C.~W. Phan, and W.~Lin, ``Enriched long-term recurrent convolutional network for facial micro-expression recognition,'' in \emph{2018 13th IEEE International Conference on Automatic Face \& Gesture Recognition (FG 2018)}.\hskip 1em plus 0.5em minus 0.4em\relax IEEE, 2018, pp. 667--674.

\bibitem{ji20123d}
S.~Ji, W.~Xu, M.~Yang, and K.~Yu, ``3d convolutional neural networks for human action recognition,'' \emph{IEEE transactions on pattern analysis and machine intelligence}, vol.~35, no.~1, pp. 221--231, 2012.

\bibitem{peng2017dual}
M.~Peng, C.~Wang, T.~Chen, G.~Liu, and X.~Fu, ``Dual temporal scale convolutional neural network for micro-expression recognition,'' \emph{Frontiers in psychology}, vol.~8, p. 1745, 2017.

\bibitem{wang2020eulerian}
Y.~Wang, H.~Ma, X.~Xing, and Z.~Pan, ``Eulerian motion based 3dcnn architecture for facial micro-expression recognition,'' in \emph{International Conference on Multimedia Modeling}.\hskip 1em plus 0.5em minus 0.4em\relax Springer, 2020, pp. 266--277.

\bibitem{xia2019spatiotemporal}
Z.~Xia, X.~Hong, X.~Gao, X.~Feng, and G.~Zhao, ``Spatiotemporal recurrent convolutional networks for recognizing spontaneous micro-expressions,'' \emph{IEEE Transactions on Multimedia}, vol.~22, no.~3, pp. 626--640, 2019.

\bibitem{sun2020dynamic}
B.~Sun, S.~Cao, D.~Li, J.~He, and L.~Yu, ``Dynamic micro-expression recognition using knowledge distillation,'' \emph{IEEE Transactions on Affective Computing}, vol.~13, no.~2, pp. 1037--1043, 2020.

\bibitem{Zhao2022MEPLANAD}
S.~Zhao, H.~Tang, S.~Liu, Y.~Zhang, H.~Wang, T.~Xu, E.~Chen, and C.~Guan, ``Me-plan: A deep prototypical learning with local attention network for dynamic micro-expression recognition,'' \emph{Neural networks : the official journal of the International Neural Network Society}, vol. 153, pp. 427--443, 2022.

\bibitem{Xie2020AUassistedGA}
H.-X. Xie, L.~Lo, H.-H. Shuai, and W.-H. Cheng, ``Au-assisted graph attention convolutional network for micro-expression recognition,'' \emph{Proceedings of the 28th ACM International Conference on Multimedia}, 2020.

\bibitem{Lei2021MicroexpressionRB}
L.~Lei, T.~Chen, S.~Li, and J.~Li, ``Micro-expression recognition based on facial graph representation learning and facial action unit fusion,'' \emph{2021 IEEE/CVF Conference on Computer Vision and Pattern Recognition Workshops (CVPRW)}, pp. 1571--1580, 2021.

\bibitem{10219571}
Y.~Zhang, H.~Wang, Y.~Xu, X.~Mao, T.~Xu, S.~Zhao, and E.~Chen, ``Adaptive graph attention network with temporal fusion for micro-expressions recognition,'' in \emph{2023 IEEE International Conference on Multimedia and Expo (ICME)}, 2023, pp. 1391--1396.

\bibitem{Hong2022LateFV}
J.~Hong, C.~Lee, and H.~Jung, ``Late fusion-based video transformer for facial micro-expression recognition,'' \emph{Applied Sciences}, 2022.

\bibitem{nguyen2023micron}
X.-B. Nguyen, C.~N. Duong, X.~Li, S.~Gauch, H.-S. Seo, and K.~Luu, ``Micron-bert: Bert-based facial micro-expression recognition,'' in \emph{Proceedings of the IEEE/CVF Conference on Computer Vision and Pattern Recognition}, 2023, pp. 1482--1492.

\bibitem{ekman1978facial}
P.~Ekman and W.~V. Friesen, ``Facial action coding system,'' \emph{Environmental Psychology \& Nonverbal Behavior}, 1978.

\bibitem{fleiss1971measuring}
J.~L. Fleiss, ``Measuring nominal scale agreement among many raters.'' \emph{Psychological bulletin}, vol.~76, no.~5, p. 378, 1971.

\bibitem{jiang2020dfew}
X.~Jiang, Y.~Zong, W.~Zheng, C.~Tang, W.~Xia, C.~Lu, and J.~Liu, ``Dfew: A large-scale database for recognizing dynamic facial expressions in the wild,'' in \emph{Proceedings of the 28th ACM International Conference on Multimedia}, 2020, pp. 2881--2889.

\bibitem{varanka2023data}
T.~Varanka, Y.~Li, W.~Peng, and G.~Zhao, ``Data leakage and evaluation issues in micro-expression analysis,'' \emph{IEEE Transactions on Affective Computing}, 2023.

\bibitem{SAN}
X.~Dong, Y.~Yan, W.~Ouyang, and Y.~Yang, ``Style aggregated network for facial landmark detection,'' \emph{2018 IEEE/CVF Conference on Computer Vision and Pattern Recognition}, pp. 379--388, 2018.

\bibitem{GPA}
J.~C. Gower, ``Generalized procrustes analysis,'' \emph{Psychometrika}, vol.~40, pp. 33--51, 1975.

\bibitem{RTF}
J.~Deng, J.~Guo, Y.~Zhou, J.~Yu, I.~Kotsia, and S.~Zafeiriou, ``Retinaface: Single-stage dense face localisation in the wild,'' \emph{ArXiv}, vol. abs/1905.00641, 2019.

\bibitem{Hara2018R3D}
K.~Hara, H.~Kataoka, and Y.~Satoh, ``Can spatiotemporal 3d cnns retrace the history of 2d cnns and imagenet?'' \emph{2018 IEEE/CVF Conference on Computer Vision and Pattern Recognition}, pp. 6546--6555, 2018.

\bibitem{Carreira2017I3D}
J.~Carreira and A.~Zisserman, ``Quo vadis, action recognition? a new model and the kinetics dataset,'' \emph{2017 IEEE Conference on Computer Vision and Pattern Recognition (CVPR)}, pp. 4724--4733, 2017.

\bibitem{Liong2018OFFApexNetOM}
S.~Liong, Y.~S. Gan, W.-C. Yau, Y.-C. Huang, and T.~Ken, ``Off-apexnet on micro-expression recognition system,'' \emph{ArXiv}, vol. abs/1805.08699, 2018.

\bibitem{Liong2019ShallowTS}
S.~Liong, Y.~S. Gan, J.~See, and H.-Q. Khor, ``Shallow triple stream three-dimensional cnn (ststnet) for micro-expression recognition,'' \emph{2019 14th IEEE International Conference on Automatic Face \& Gesture Recognition (FG 2019)}, pp. 1--5, 2019.

\bibitem{Xia2020RevealingTI}
Z.~Xia, W.~Peng, H.-Q. Khor, X.~Feng, and G.~Zhao, ``Revealing the invisible with model and data shrinking for composite-database micro-expression recognition,'' \emph{IEEE Transactions on Image Processing}, vol.~29, pp. 8590--8605, 2020.

\end{thebibliography}




%

\begin{IEEEbiography}[{\includegraphics[width=1in,height=1.25in,clip,keepaspectratio]{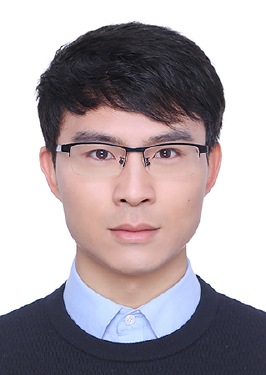}}]{Sirui Zhao} received the PhD
degree with the Department of Computer Science
and Technology from University of Science and
Technology of China (USTC).  He is also a faculty member with the Southwest University of Science and Technology. His research interests
include automatic micro-expressions analysis,
human-computer interaction (HCI) and affect computing. He has published several papers in refereed
conferences and journals, including ACM Multimedia Conference, ICME, IEEE Transactions on Affective Computing, ACM TOMM, Neural Networks, etc.
\end{IEEEbiography}

\begin{IEEEbiography}[{\includegraphics[width=1in,height=1.25in,clip,keepaspectratio]{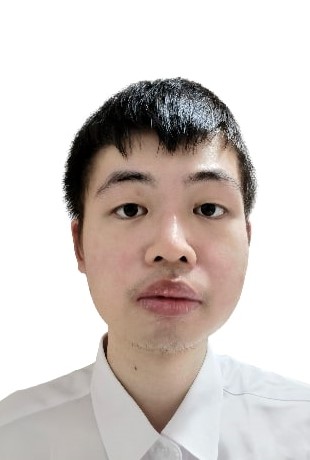}}]{Huaying Tang}
received the B.S. degree in the School of Computer Science and Technology from University of Science and Technology of China (USTC), Hefei, China, in 2021. He is currently pursuing the M.S. degree
in computer science and technology in USTC. His research interests
lie around automatic micro-expressions analysis and affect computing. He has published several papers in refereed conferences and journals, including ACM Multimedia, Neural Networks, etc.
\end{IEEEbiography}

\begin{IEEEbiography}[{\includegraphics[width=1in,height=1.25in,clip,keepaspectratio]{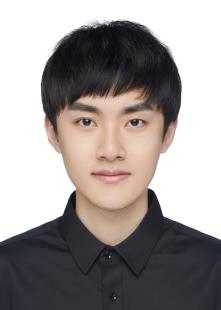}}]{Xinglong Mao} received the B.S degree in the School of Data Science from University of Science and Technology of China (USTC), Hefei, China. He is currently working toward the M.S. degree from the School of Data Science. His research interests
include automatic micro-expressions analysis and affect computing. He has published several conference papers in ACM Multimedia Conference, ICME, etc.
\end{IEEEbiography}

\begin{IEEEbiography}[{\includegraphics[width=1in,height=1.25in,clip,keepaspectratio]{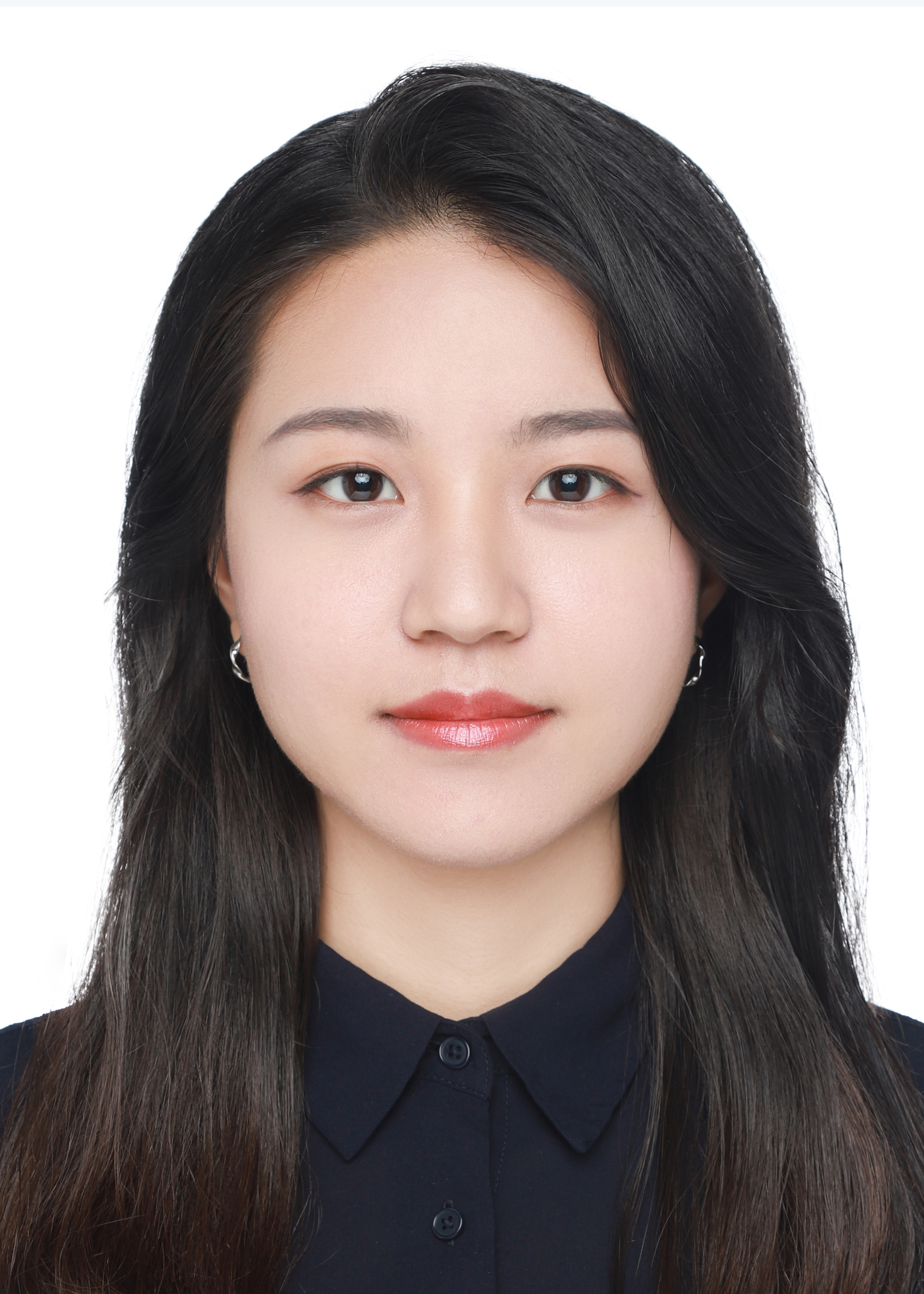}}]{Shifeng Liu} received the B.S degree in the School of Gifted Young from University of Science and Technology of China (USTC), Hefei, China. She is currently working toward the M.S. degree from the School of Data Science. Her research interests
include automatic micro-expressions analysis,
human-computer interaction (HCI) and affect computing. She has published several papers in refereed
conferences and journals, including ACM Multimedia Conference, ICME, Neural Networks, etc.
\end{IEEEbiography}

\begin{IEEEbiography}[{\includegraphics[width=1in,height=1.25in,clip,keepaspectratio]{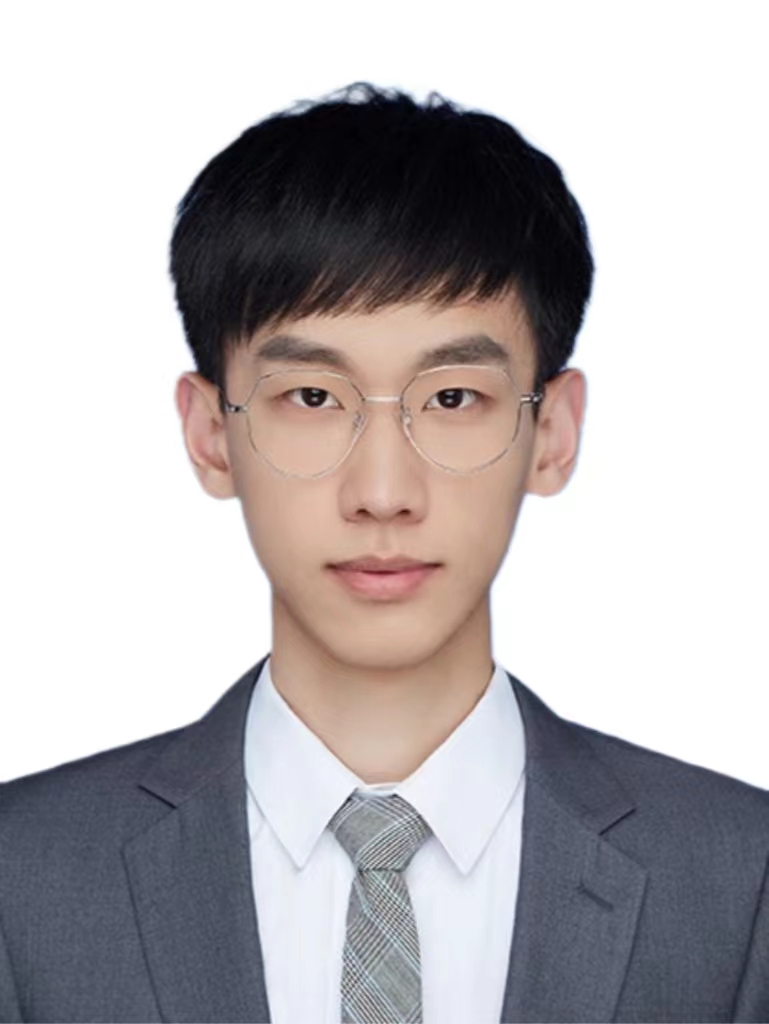}}]{Yiming Zhang} is currently working toward the Ph.D. degree in the School of Data Science from University of Science and Technology of China (USTC). His research interests include automatic micro-expressions analysis and affect computing. He has published several papers in conference proceedings, including ACM Multimedia and ICME.
\end{IEEEbiography}

\begin{IEEEbiography}[{\includegraphics[width=1in,height=1.25in,clip,keepaspectratio]{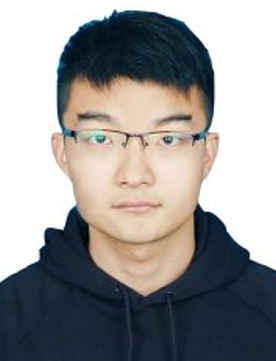}}]{Hao Wang} received the PhD degree in computer science from USTC. He is currently an associate researcher with the School of Computer Science
and Technology, USTC. His main research interests include data mining, representation learning, network embedding and recommender systems. He has published several papers in referred conference proceedings, such as TKDE, TOIS, NeuriPS, and AAAI.
\end{IEEEbiography}

\begin{IEEEbiography}[{\includegraphics[width=1in,height=1.25in,clip,keepaspectratio]{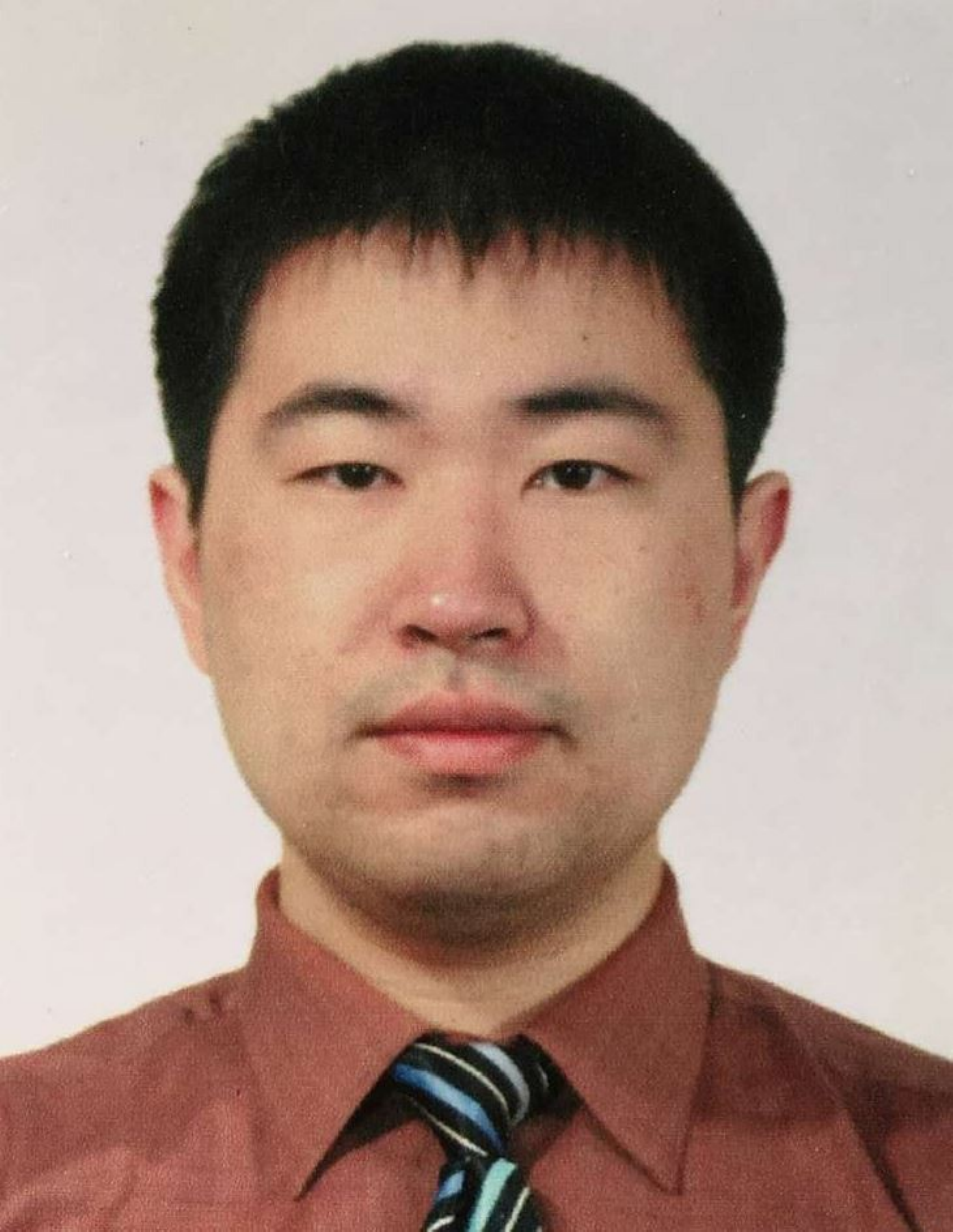}}]{Tong Xu}
received the Ph.D. degree in University of Science and Technology of China (USTC), Hefei, China, in 2016. He is currently working as a Professor of the Anhui Province Key Laboratory of Big Data Analysis and Application, USTC. He has authored 100+ journal and conference papers in the fields of social network and social media analysis, including IEEE TKDE, IEEE TMC, IEEE TMM, KDD, AAAI, ICDM, etc.
\end{IEEEbiography}

\begin{IEEEbiography}[{\includegraphics[width=1in,height=1.25in,clip,keepaspectratio]{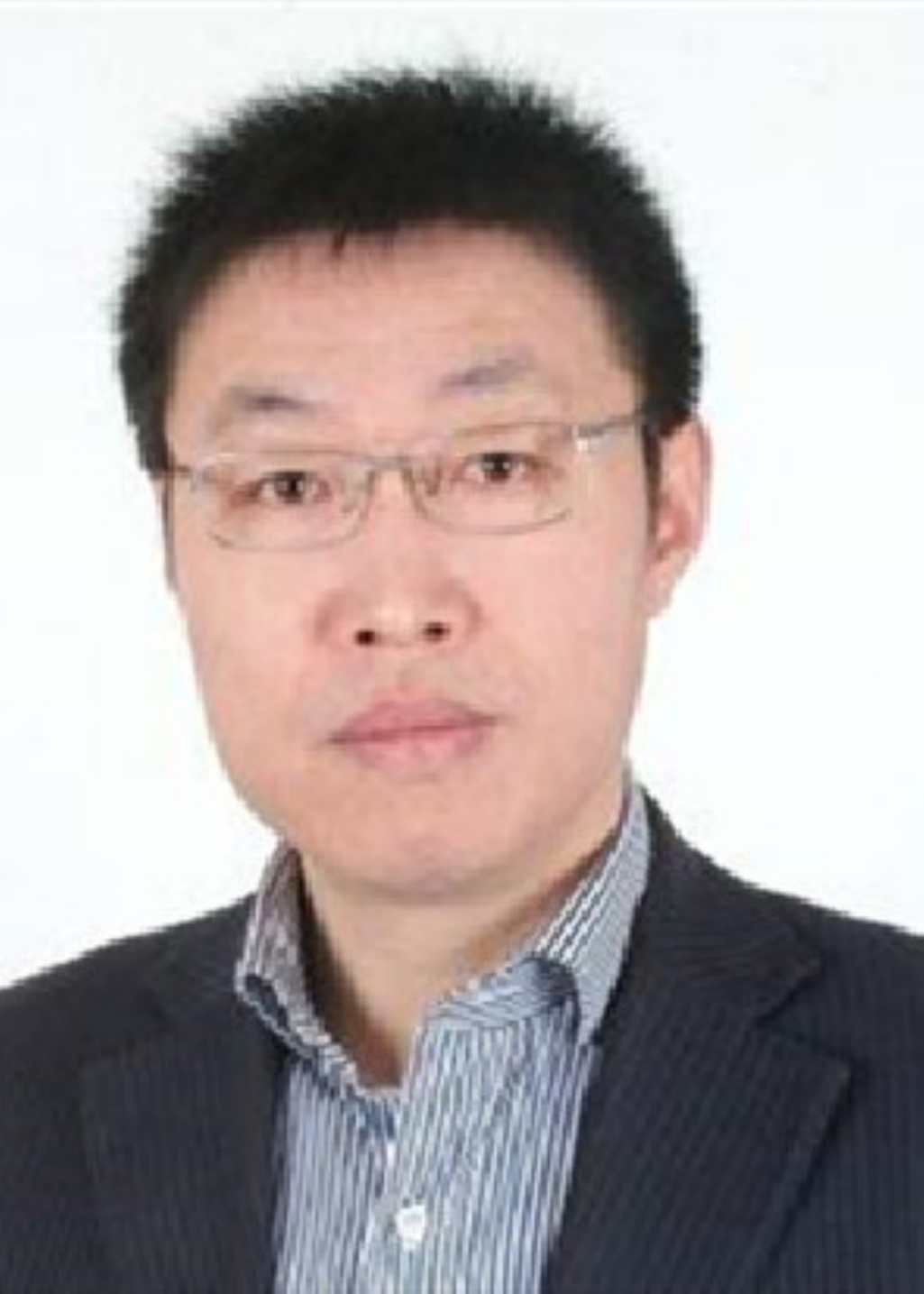}}]{Enhong Chen} (Fellow, IEEE) received the PhD degree from USTC. He is a professor and executive dean of School of Data Science, USTC.
His general area of research includes data mining
and machine learning, social network analysis, and
recommender systems. He has published more
than 200 papers in refereed conferences and journals, including IEEE Transactions on Knowledge
and Data Engineering, IEEE Transactions on Mobile
Computing, KDD, ICDM, NeurIPS, and CIKM. He
was on program committees of numerous conferences including KDD, ICDM, and SDM. His research is supported by the
National Science Foundation for Distinguished Young Scholars of China.
\end{IEEEbiography}





\end{document}